\pdfoutput=1

\documentclass[11pt]{article}

\usepackage[final]{acl}

\usepackage{times}
\usepackage{latexsym}

\usepackage[T1]{fontenc}

\usepackage[utf8]{inputenc}

\usepackage{microtype}

\usepackage{inconsolata}

\usepackage{graphicx}

\usepackage{amsmath}    
\usepackage{booktabs}
\usepackage{pifont}
\usepackage{enumitem}
\usepackage[utf8]{inputenc}  
\usepackage[T1]{fontenc} 
\usepackage{arydshln}   

\usepackage[most]{tcolorbox}
\newtcolorbox{mybox}[1][]{%
  colback=blue!5!white,    
  colframe=blue!75!black,  
  fonttitle=\bfseries,
  title=#1,
  boxrule=0.8pt,           
  arc=2mm,                 
  left=2mm,
  right=2mm,
  top=1mm,
  bottom=1mm
}
\definecolor{my-blue}{HTML}{4B9CD3}
\usepackage{setspace}

\newcommand{\UnreliableNarrator}{\textsc{TUNa}}

\usepackage{ulem}

\title{Classifying Unreliable Narrators with Large Language Models}

\author{
  Anneliese Brei$^1$\thanks{Department of Computer Science} \quad 
  Katharine Henry$^1$\thanks{Department of English and Comparative Literature} \quad
  Abhisheik Sharma$^{2*}$  \quad \\
  \textbf{Shashank Srivastava$^{1*}$} \quad   
  \textbf{Snigdha Chaturvedi$^{1*}$}  \\
  $^1$UNC Chapel Hill  \quad $^2$Virginia Polytechnic Institute and State University \\
  \texttt{\href{mailto:abrei@cs.unc.edu}{abrei@cs.unc.edu}},\quad
  \texttt{\href{mailto:katharinehenry@alumni.unc.edu}{katharinehenry@alumni.unc.edu}},\quad
  \texttt{\href{mailto:abhisharma@vt.edu}{abhisharma@vt.edu}},\quad\\
  \texttt{\href{mailto:ssrivastava@cs.unc.edu, snigdha@cs.unc.edu}{\{ssrivastava, snigdha\} @cs.unc.edu}}\\
  }

\begin{document}
\maketitle
\begin{abstract}

Often when we interact with a first-person account of events, we consider whether or not the narrator, the primary speaker of the text, is reliable. In this paper, we propose using computational methods to identify unreliable narrators, i.e. those who unintentionally misrepresent information. Borrowing literary theory from narratology to define different types of unreliable narrators based on a variety of textual phenomena, we present \UnreliableNarrator, a human-annotated dataset of narratives from multiple domains, including blog posts, subreddit posts, hotel reviews, and works of literature. We define classification tasks for intra-narrational, inter-narrational, and inter-textual unreliabilities and analyze the performance of popular open-weight and proprietary LLMs for each. We propose learning from literature to perform unreliable narrator classification on real-world text data. To this end, we experiment with few-shot, fine-tuning, and curriculum learning settings. Our results show that this task is very challenging, and there is potential for using LLMs to identify unreliable narrators. We release our expert-annotated dataset and code at
\href{https://github.com/adbrei/unreliable-narrators}{https://github.com/adbrei/unreliable-narrators}
and invite future research in this area.
\end{abstract}

\section{Introduction}
Imagine that you are on social media warning your friends about a recent shopping experience, and before submitting the post, you wonder if the presentation of your writing undermines your credibility.
In another window, you are writing a cover letter. You recount a critical learning experience from your past job and wonder if your present voice sounds reliable to the reader. In the next room, your family is discussing the debate transcript between political candidates. Your sister thinks one of the candidates speaks like a villain from a novel she read, and your family differs on how reliable the candidate actually is. For each of these situations, it would be useful to have an automatic tool that identifies unreliability.

\begin{figure}
    \centering
    \includegraphics[width=\linewidth]{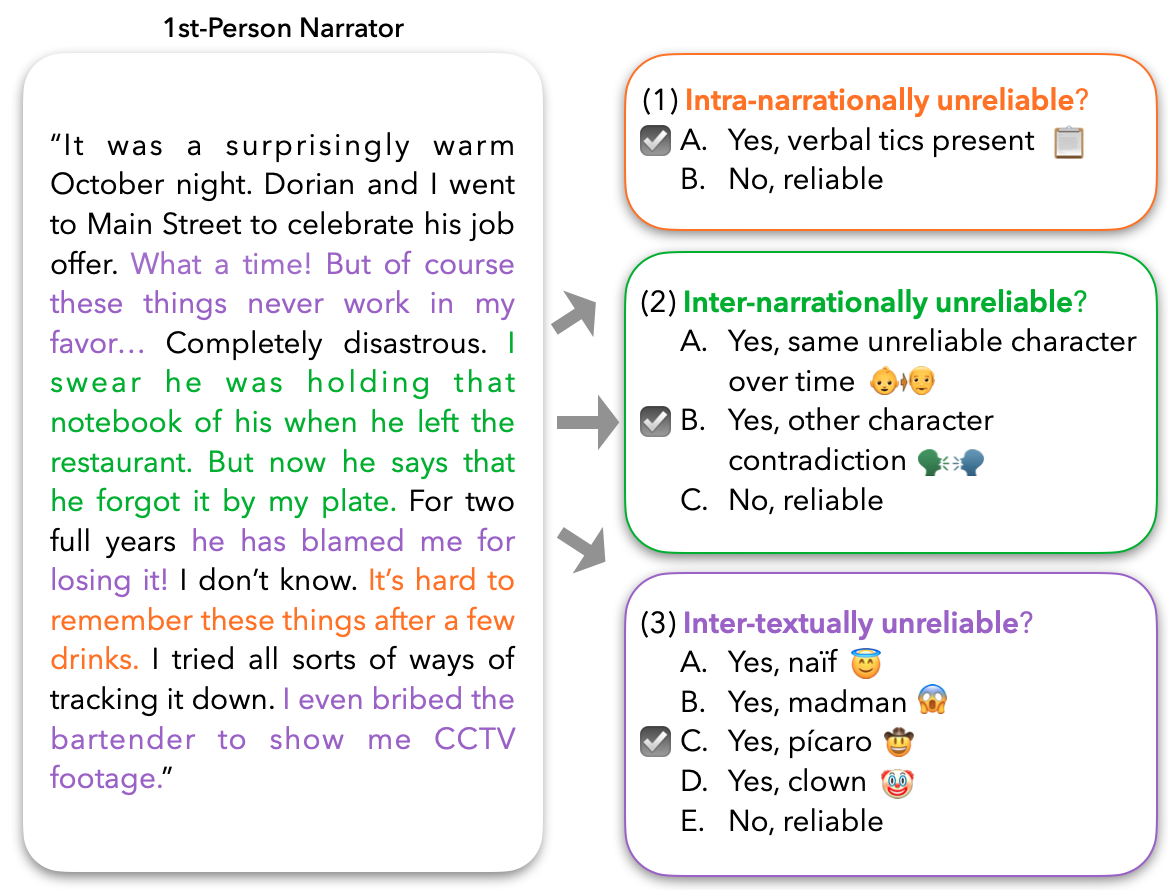}
    \caption{Real-world text with first-person narrators, such as the narrative shown (left), can be analyzed to determine the unreliability of the narrator. We separately classify three types of unreliability (right): intra-narrational, inter-narrational, and inter-textual. }
    \label{fig:narrative-example}
\end{figure}

Readers of personal accounts, such as reviews, online comments, cover letters, and college application essays, often implicitly question the reliability of the narrator: \textit{Can I trust how this narrator has perceived and is describing the event?}
Meanwhile, writers who wish to defend their points are concerned about how they textualize their ideas: \textit{Am I sharing information in a reliable way?} Answering such questions is critical for the safe transmission of information \cite{nunning2015conceptualising}.

However, answering these questions is not a simple task.
That is because unreliability cues are often subtle and context-dependent \cite{hansen2007reconsidering}. They might be scattered across the text or involve a deeper understanding beyond what is explicitly stated. Sometimes it is necessary to draw abstract inferences about the emotional and mental state of the narrator. Also, a text might have many readers, some of whom focus on different aspects of these cues. From a writer's perspective, it is important to pay attention to all of these cues to ensure the writing sounds reliable to all readers.

Narratology has explored these questions by attempting to define the \textit{unreliable narrator}, \textbf{a first-person speaker who \textit{unintentionally} describes situations misleadingly}  \cite{booth1983rhetoric}. \citet{hansen2007reconsidering} considers leading definitions and observes ``the unreliable narrator is a concept covering very diverse textual phenomena'' and accordingly proposes a taxonomy containing different forms of unreliable narrators with ``conceptual distinction.''

These forms include intra-narrational, inter-narrational, and inter-textual unreliability. The first form, \textbf{\textit{intra-narrational unreliability}} is the classical definition that focuses on the presence of verbal tics (textual cues that indicate uncertainty). The left half of Figure~\ref{fig:narrative-example} shows an example: an excerpt from a blog post where the writer narrates an experience with another person, Dorian, at a bar. The text in orange font indicates content that is narrated in an intra-narrationally unreliable manner because the narrator admits having trouble remembering details due to being inebriated. Consequently their narration is possibly unreliable.
The second form, \textbf{\textit{inter-narrational unreliability}} occurs when a secondary voice presents a contrasting version of events. For example, in Figure~\ref{fig:narrative-example}, highlighted in green, Dorian does not agree with the narrator regarding the whereabouts of a notebook. Such a contradiction indicates that either the narrator or Dorian must be wrong and raises reader's doubts about the reliability of the narrator.  
The third form, \textbf{\textit{inter-textual unreliability}} involves pattern-matching the narrator with established unreliable character tropes \cite{riggan1978picaros}. In Figure~\ref{fig:narrative-example}, highlighted in purple, the reader questions the narrator's reliability because they seem cunning as they bribe the bartender (fitting the trope of \textit{pícaro}). More detailed definitions of each type of unreliability are outlined in Section \ref{sec:definitions-of-unreliability}. 

Identifying these three forms of unreliable narrators requires picking up on subtle cues that range from specific lexical choices (e.g., a direct statement such as ``it's hard to remember'') to increasingly abstract inferences (e.g., drawing inferences through statements and actions that a narrator has cunning and self-interest). These forms may contain overlapping characteristics; however, they are classified and determined separately. Hence, a narrator might be unreliable in one of these forms but not another. It is valuable to analyze narrators in this way because it provides in-depth views of the narrator from lexical to abstract contextual levels. Determining narrator unreliability ultimately considers all three forms since together they provide a more complete picture.

In this work, we borrow these definitions from narratology and introduce the task of automatically identifying three forms of unreliable narrators. We pose this problem as a set of binary/multi-class classifications corresponding to the three types of unreliability (shown in the right of Figure~\ref{fig:narrative-example}). 
We propose that these ideas from the theoretical field of narratology can be used more broadly to identify unreliability across diverse real-world domains. 

We observe that as of date there has been no work on analyzing narrator unreliability with automatic methods, and there are no available resources or labeled datasets. Hence, we introduce \UnreliableNarrator, a collection of personal anecdotes from blog posts, subreddit posts, online reviews, and fiction. We hire expert annotators obtaining honors undergraduate or graduate degrees in English literature to annotate these accounts for the three forms of unreliability mentioned above.

To identify unreliable narrators automatically, we explore using large language models (LLMs). We conduct experiments with 6 open and closed-source LLMs of a variety of sizes. We try zero/few-shot settings, fine-tuning, and curriculum learning \cite{bengio2009curriculum}. With these methods, we attempt to learn from labeled data from fiction and generalize this knowledge to real-world text. We observe that classifying unreliable narrators is a very difficult task and encourage future research to further explore its nuances and challenges. Our contributions are as follows:

\begin{itemize}[topsep=1pt, leftmargin=*, noitemsep]
    \item We introduce the task of automatically identifying unreliable narrators;
    
    \item We borrow narratological definitions for unreliable narrator (i.e., we consider three diverse and increasingly abstract forms: intra-narrational, inter-narrational, and inter-textual);
    
    \item We introduce \UnreliableNarrator, an expert annotated dataset of unreliable first-person accounts spanning four different text domains;
    
    \item We experiment with multiple methods that learn how to identify unreliable narrators in snippets from fiction and transfer this knowledge to common text read in everyday situations.
    
\end{itemize}

\section{Background and Related Work} 

The term ``unreliable narrator'' is originally defined by \citet{booth1983rhetoric}: ``For lack of better terms, I have called a narrator \textit{reliable} when he speaks for or acts in accordance with the norms of the work (which is to say, the implied author's norms), \textit{unreliable} when he does not.'' The vagueness of this definition has encouraged more recent narratologists to attempt to define the unreliable narrator in more certain terms \cite{cannings2023reading, jacke2018unreliability, heyd2006understanding, olson2003reconsidering, fludernik2000unreliable, currie1995unreliability, riggan1978picaros}. \citet{culler1977very} states, ``Narrators are sometimes termed unreliable when they provide enough information about situations and clues about their own biases to make us doubt their interpretations of events...'' \citet{hansen2007reconsidering} builds upon the work of Culler and other salient narratologists to propose a taxonomy with definitions of multiple aspects of narrator unreliability.  In this work, we adopt these definitions and taxonomy. 

To the best of our knowledge, \textit{there is currently no existing literature that explores automated approaches for identifying unreliable narrators.} We note recent efforts to automatically understand other aspects of protagonists, who are sometimes depicted in first-person \cite{yuan2024evaluating,jang2024evaluating,brahman2021let,huang2021uncovering,bamman2013learning}. Additionally, some works attempt to analyze the emotions of protagonists \cite{brahman2020modeling,rahimtoroghi2017modelling} or their relationships with other characters \cite{vijjini2022towards,kim2019frowning,chaturvedi2017unsupervised,iyyer2016feuding,srivastava2016inferring}. Such works indicate that using automatic methods is a reasonable approach for addressing our task. 

Classifying unreliable narrators is to a limited extent related to tasks such as the automatic identification of misinformation \cite{saeidnia2025artificial,jarrahi2023evaluating} and rumors \cite{he2025survey,kwao2025survey}. It is also distantly related to deception detection, defined by \citet{burgoon1994interpersonal} as the identification of narrators who intend to commit \textit{deception}, ``a deliberate act perpetrated by a sender to engender in a receiver beliefs contrary to what the sender believes is true to put the receiver at a disadvantage'' \cite{hazra-majumder-2024-tell, constancio2023deception, sarzynska2023truth, fornaciari2021bertective, van2018identity, eloff2015big,almela2013seeing}. These tasks are similar because they analyze first-person narrators and consider aspects of narrative believability. However, \citet{booth1983rhetoric} draws a clear distinction by determining that conscious lying is not a characteristic of unreliable narrators; instead unreliability is ``a matter of [what is called] \textit{inconscience}; the narrator is mistaken, or he believes himself to have qualities which the author denies him.'' 
We follow Booth's reasoning and do not consider narrators who deliberately intend to mislead readers but only those who sound unreliable.

\section{Definitions of Unreliability}
\label{sec:definitions-of-unreliability}

\begin{table*}[ht]
    \centering
    \footnotesize
    \begin{tabular}{p{9cm}p{6cm}}
    \toprule
    \textbf{Type} & \textbf{Example}\\
    \midrule
    \textit{Admission of fault or bias}: Explicit admission of mistakes, biases, missing details, or reporting details from another likely unreliable character. & ``I tend to see things from a unique point of view.'', ``Like others of my generation...''  \\ 
    \midrule
    \textit{Defensive tone}: Multiple phrases in protestation. &``I feel I should explain''\\ 
    \midrule
    \textit{Digressions}: Statement that veers off-topic. & ``I will do that in a minute. By the way...''\\ 
    \midrule
    \textit{Hedging language}: Multiple phrases that indicate uncertainty or vagueness. &``it seems that'', ``it appears to be'', ``I think'', ``maybe'', ``sort of''\\ 
    \midrule
    \textit{Inconsistencies}: Two or more contradicting statements or events that do not add up.& ``I am a nobody. But look! There is a plane drawing my name in the sky.''\\ 
    \midrule
    \textit{Selective memory}: Explicit admission that narrator may have forgotten details. &``It was so long ago, it's hard to remember'', ``My memory is not what it used to be''\\ 
    \midrule
    \textit{Statement of potential disbelief}: Explicit admission that narrative sounds unlikely. &``You might not believe me, but...'', ``what happened next might seem strange''\\ 
    \bottomrule
    \end{tabular}
    \caption{Examples of verbal tics exhibited by intra-narrationally unreliable narrators.}
    \label{tab:verbal-tics}
\end{table*}

In choosing our definitions of unreliability, we make two assumptions. Firstly, given a text, we assume that it contains explicit or implicit information that can be leveraged to ascertain the narrator's unreliability \cite{chatman1990coming}. By \textit{explicit} information, we refer to statements that directly state that an account may be unreliable (e.g., the narrator admits to being inebriated during the time of the described events, as demonstrated in the narrative in Figure~\ref{fig:narrative-example}). By \textit{implicit} information, we refer to less direct details (e.g., patterns exhibited by the narrator that resemble unreliable character tropes). Secondly, following \citet{wall1994remains}, we assume a narrator is reliable until the reader notices explicit or implicit information indicating unreliability.

We borrow definitions from the taxonomy for unreliable narration introduced by \citet{hansen2007reconsidering}. We note this taxonomy, proposed as the culmination of a broad range of prior definitions, provides a diverse set of tools for analyzing narrators from different perspectives and levels of difficulty. We use three forms that analyze traits with increasingly abstract conceptions of unreliability, as described in the next three subsections. Examples for each of these unreliable forms from the different textual domains are given in Appendix \ref{sec:unreliability-types-examples}.

\subsection{Intra-narrational Unreliability}
In intra-narrational unreliability the narrator exhibits verbal tics, ``small interjections and comments that hint at an uncertainty in the narrator's relating of the events'', such as ``I think'' or ``it was so long ago, it's hard to remember.'' \cite{hansen2007reconsidering}. Table \ref{tab:verbal-tics} shows various types of verbal tics and corresponding examples. If at least one type of verbal tic is present in a text, its narrator is considered intra-narrationally unreliable.

\subsection{Inter-narrational Unreliability}
In this form, the narrator is unreliable from a secondary point of view as in the following two cases: 

\textbf{Same-unreliable-character-over-time} \vspace{-1ex}\includegraphics[height=1em]{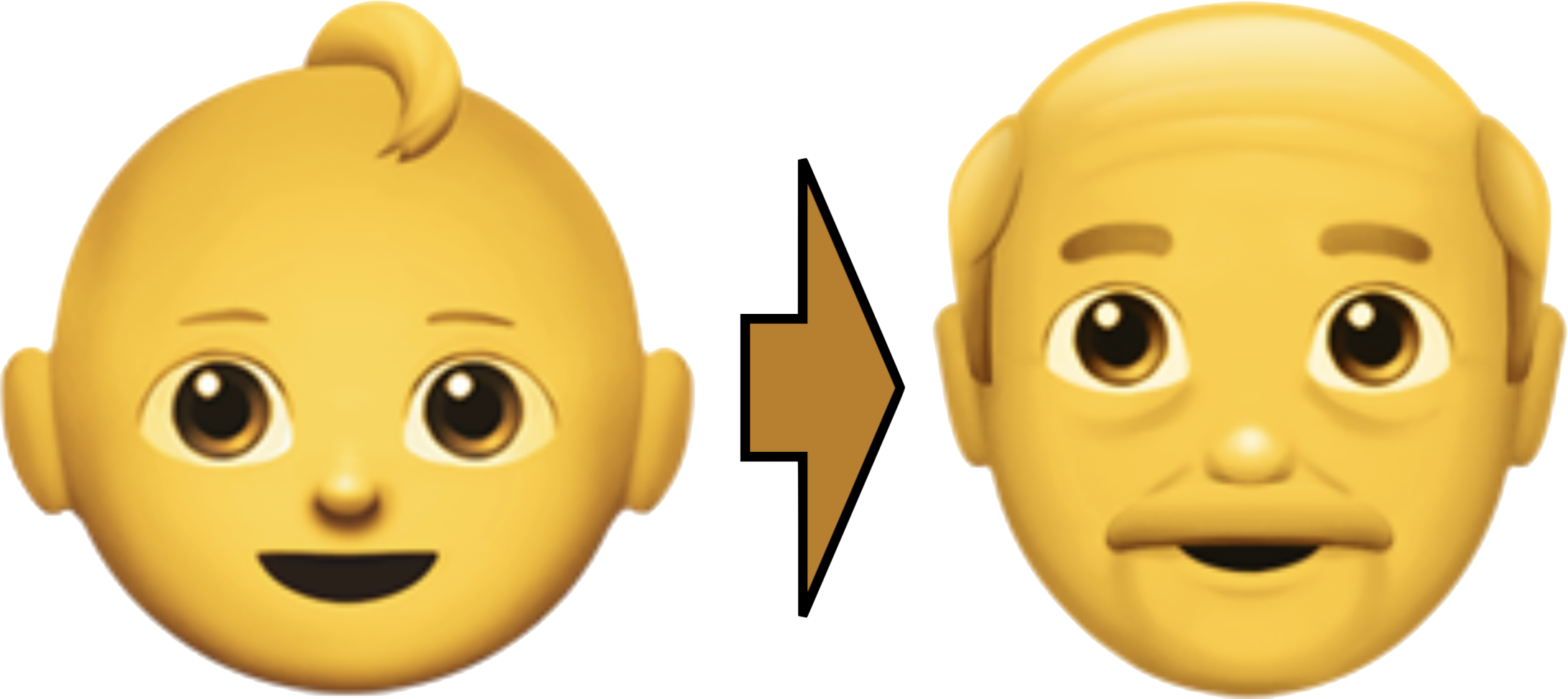}\vspace{1ex}: The narrator is reflecting on events in the distant past when he/she exhibits traits of unreliability \textit{and} the present-day narrator does not indicate change within the narrative snippet (i.e., the current voice of the narrator has traits of unreliability). For example: \textit{``I used to be a crazy man. I’d wait in line each day, desperately hoping that they would let me in. Weee, those were good times.''} In this snippet, the narrator describes his distant past as unreliable with ``I used to be a crazy man...'' His last statement, ``Weee, those were good times'', indicates his perspective has not changed over time.

\textbf{Other-character-contradiction} \vspace{-1ex}\includegraphics[height=1em]{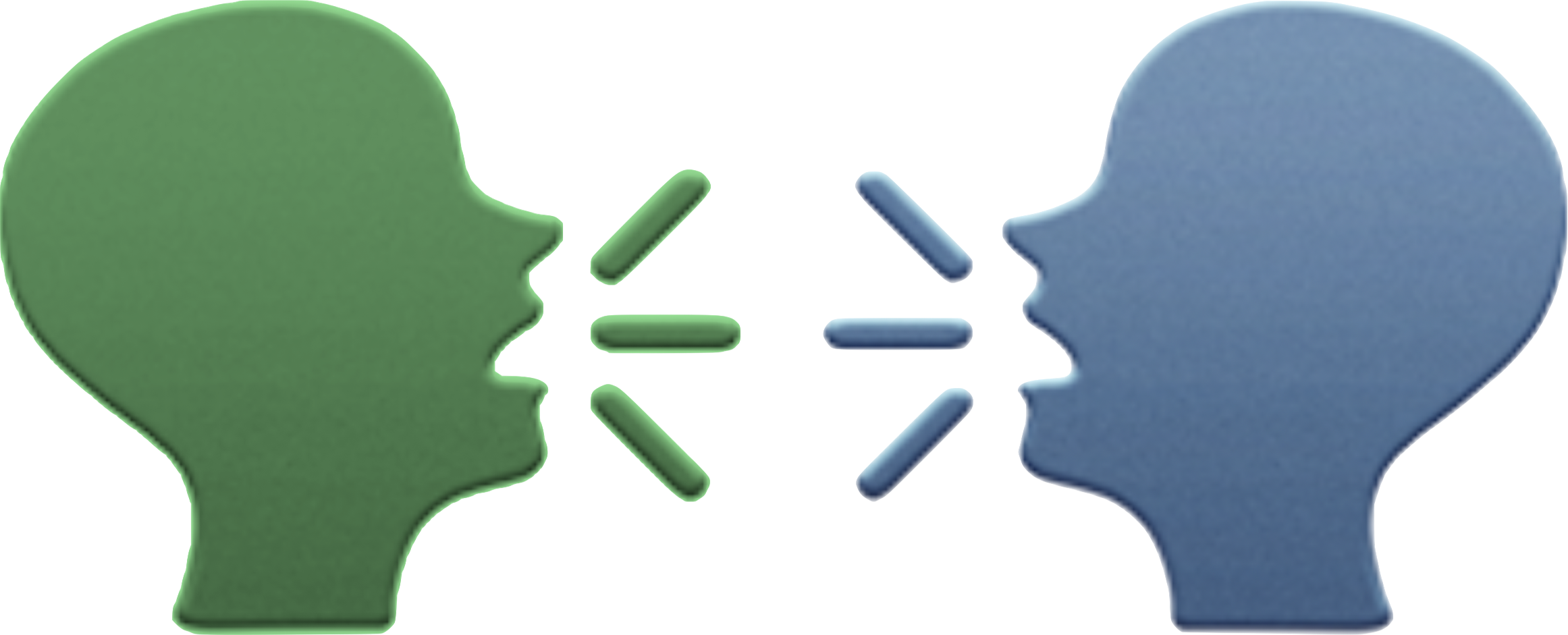}\vspace{1ex}: Another character contradicts the narrator, typically in the form of direct dialogue. For example: \textit{``I thought the offer from Henry’s was incredible. As I picked up a pen to sign, I heard the judge’s voice: he had entered the room through the far door and was talking to two well-dressed men. “What scammers these men from Henry’s have become,” he was saying.''} In this snippet, the narrator believes he has received a good offer, but another character, a judge, has a contradicting perspective that the offer is a scam. The reader does not know which character understands the situation best, leaving the narrator's reliability in doubt.

\subsection{Inter-textual Unreliability}
In this form, if the narrator fits the description of one of the following unreliable character tropes, as defined by \citet{riggan1978picaros}, the narrator is considered inter-textually unreliable:

\textbf{Naïf} \vspace{-1ex}\includegraphics[height=1em]{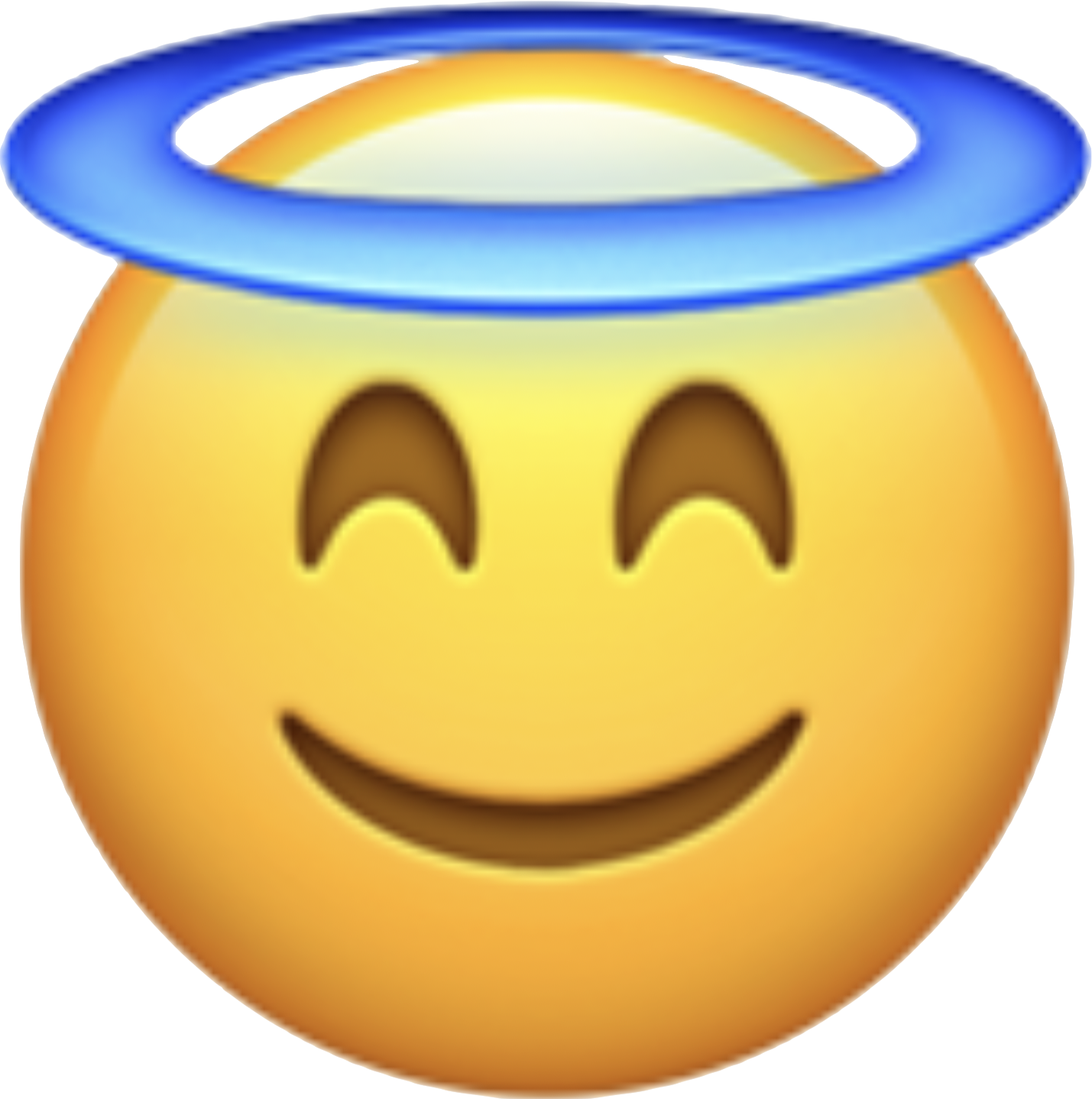}\vspace{1ex}: \textit{Blind to wrongs.} Naive observer who lacks the social savvy, maturity, or awareness to understand the complexity of their environment. For example: \textit{``I accepted the assignment willingly. Dimly, I heard the people around me muttering – talking about some danger? I ignored them and went to the other room.''} In this snippet, the narrator acts blindly without understanding the situation.
        
\textbf{Madman} \vspace{-1ex}\includegraphics[height=1em]{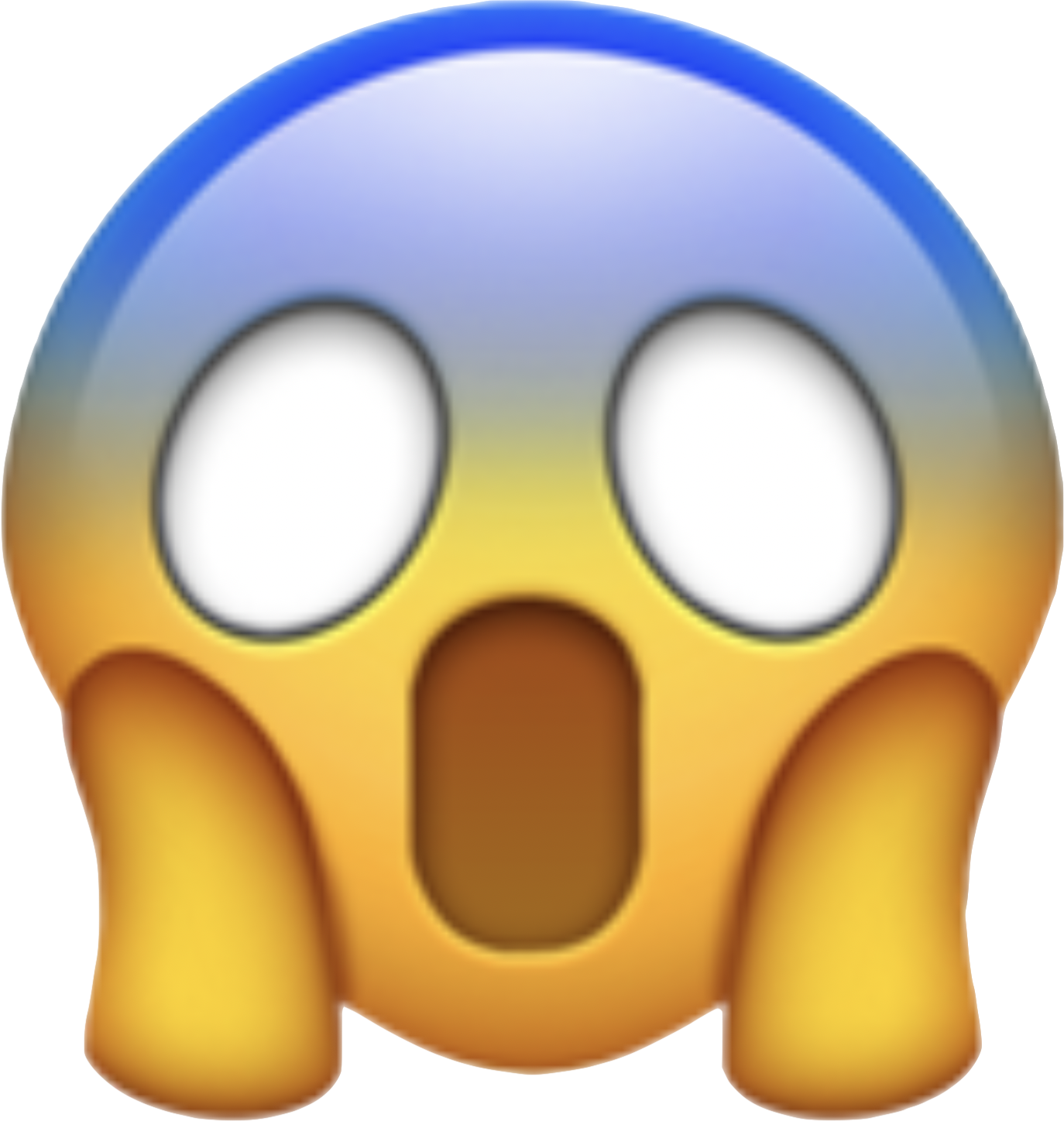} \vspace{1ex}: \textit{Highly emotional.} Narrator, often with a frantic voice, who feels deep positive or negative emotions toward others and is maddened by perceived torture or alienation. For example: \textit{``My heart beat wildly. It took my greatest strength to turn and walk away. How could he? My best friend, a betrayer?!''} In this snippet, the narrator reveals deep negative feelings, perceived alienation, and a frantic tone revealed through stylistic choices.
        
\textbf{Pícaro} \vspace{-1ex}\includegraphics[height=1em]{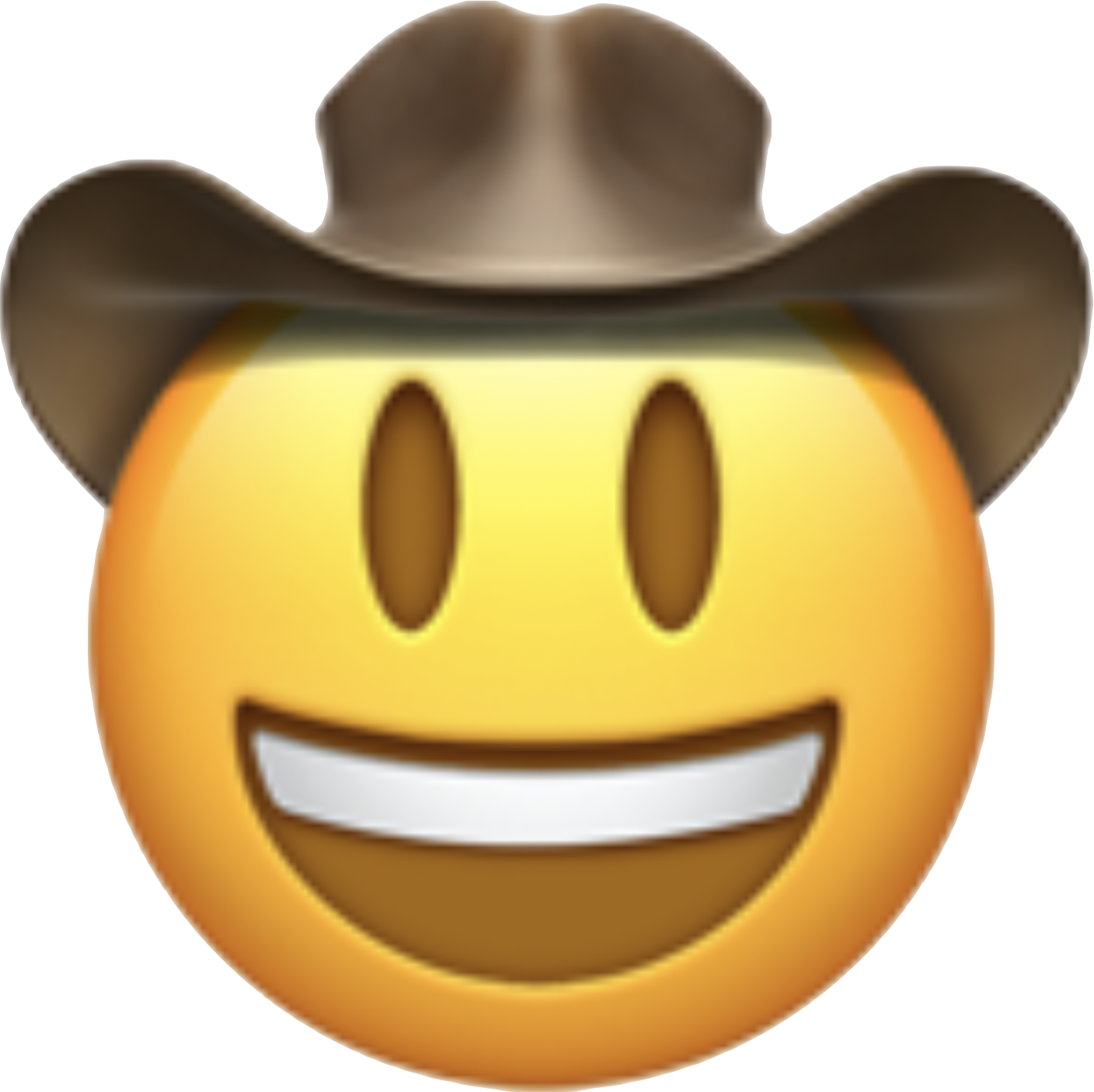}\vspace{1ex}: \textit{Tries to be cunning.}
Socially aware rogue or antihero who experiences the rise and fall of fortune while attempting to improve their prospects and cleverly justifying their chaotic worldview. For example: \textit{``The school teacher scolded me and took away the paper airplane. As soon as her back was turned, I whipped out a fresh sheet of paper, determined to be more stealthy this time. All the while, I kept one eye on the girl who had reported me.''} This narrator experiences a fall of fortune when his paper airplane is taken away. He tries to improve his prospects by making a new airplane and shows cunning when he stealthily tries to avoid being caught again.
        
\textbf{Clown} \vspace{-1ex}\includegraphics[height=1em]{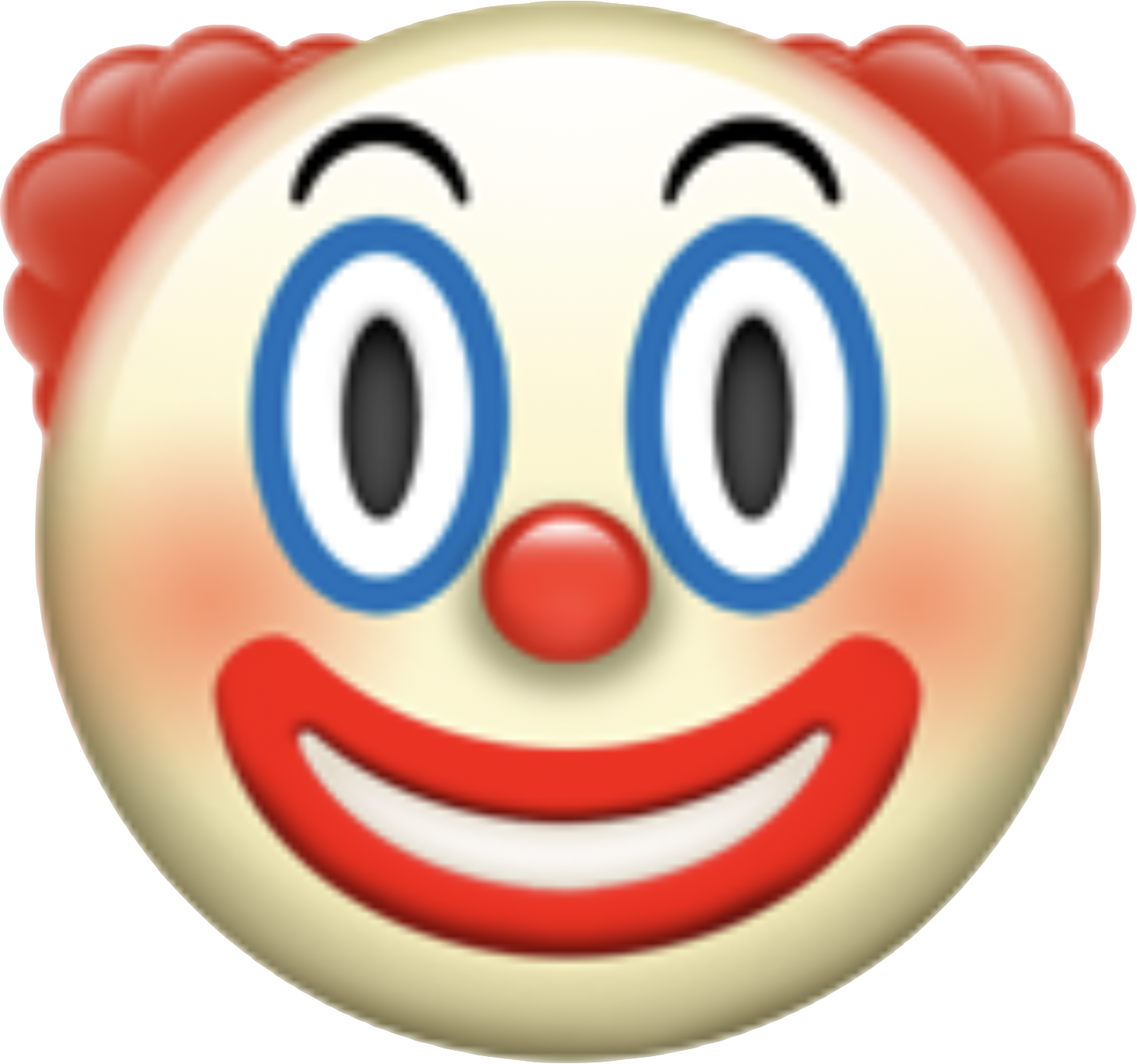}\vspace{1ex}: \textit{Flips the narrative.} Narrator who offers reinterpretations that repackage internal and/or external conflict in a new light, potentially from behind a facade that allows them to say whatever they want. For example: \textit{``They called me a coward. What ho! I saw myself rather as my own liberator.''} This narrator describes a societal view (that they are a coward) and makes it clear they have a different, reinterpreted view (that they are a liberator).

\subsection{The \UnreliableNarrator\ Dataset} 

\begin{table}
  \centering
  \footnotesize
  \begin{tabular}{lcccc}
    \toprule
    \textbf{Corpus} & \textbf{\# Samples} & \textbf{Avg} & \textbf{Min} & \textbf{Max} \\
    \toprule
    Fiction     & 499 & 194.31 & 24 & 924 \\
    \hspace{.25cm}\textit{Train/Valid}      & 373 & 194.74 & 24 & 514\\
    \hspace{.25cm}\textit{Test}             & 126 & 193.06 & 48 & 924\\
    Blog posts  & 106 & 315.31 & 114 & 1050 \\
    Subreddit   & 112 & 396.88 & 73 & 858 \\
    Reviews     & 100 & 157.43 & 53 & 460 \\
    \midrule
  \end{tabular}
  \caption{\UnreliableNarrator\ statistics, including the total number of samples and the average, minimum, and maximum number of tokens in each sample per domain. The first row of Fiction is the combination of \textit{Train/Valid} and \textit{Test} subsets (rows 2 and 3 respectively).}
  \label{tab:corpora-stats}
\end{table}

\begin{figure}[t]
    \centering
    \includegraphics[width=\linewidth, height=120pt]{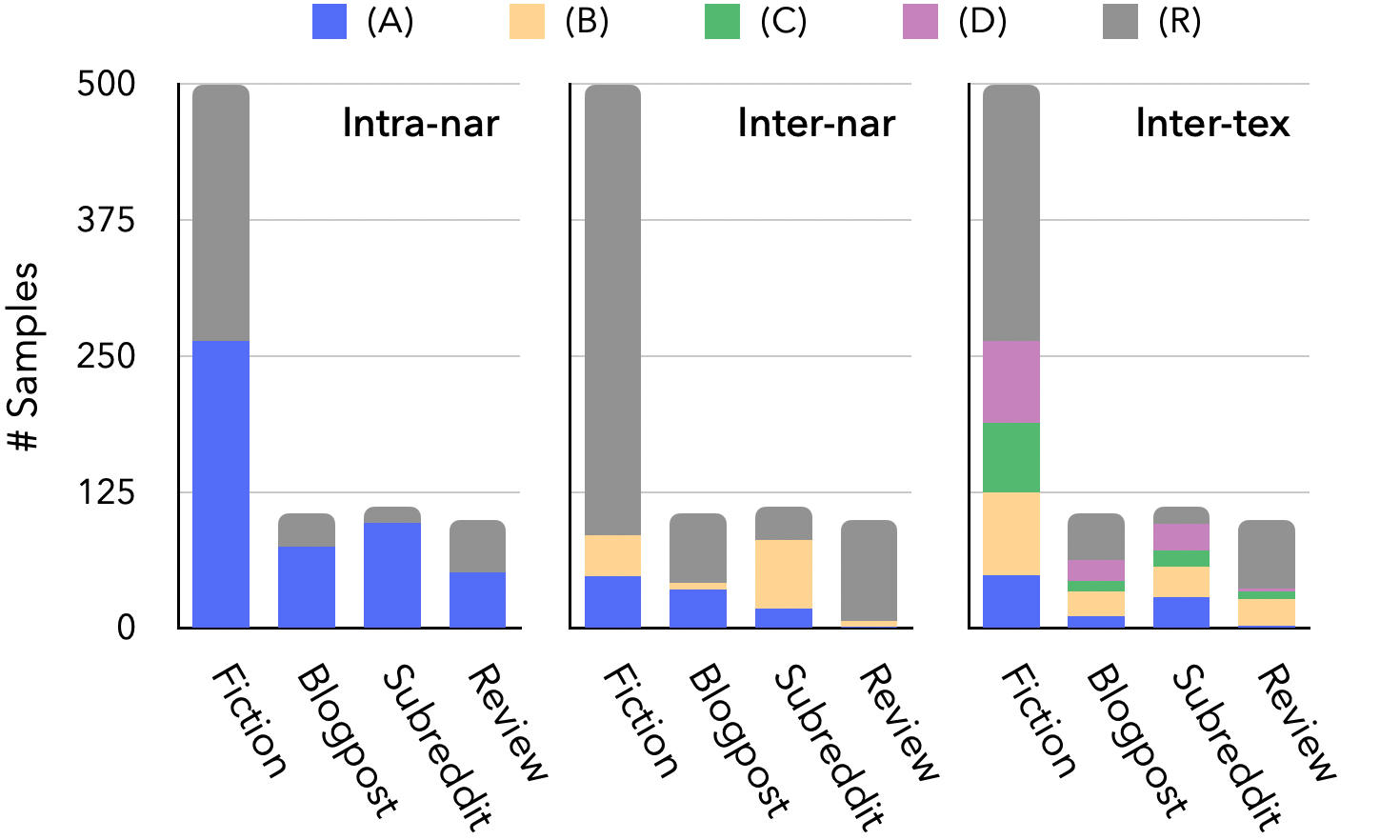}
    \caption{Distribution of resolved labels. For intra-nar (left): \# narratives with \textbf{(A)} verbal tics or \textbf{(R)} none (reliable). For inter-nar (middle): \# narratives with \textbf{(A)} ``same unreliable character over time'', \textbf{(B)} ``other character contradiction'', or \textbf{(R)} none. For inter-tex (right): \# narratives with \textbf{(A)} naïf, \textbf{(B)} madman, \textbf{(C)} pícaro, \textbf{(D)} clown, or \textbf{(R)} none.}
    \label{fig:label-distribution}
\end{figure}

Since there are no currently available resources for classifying unreliable narrators, we build an expert-annotated dataset, \textbf{T}exts with \textbf{U}nreliable \textbf{Na}rrators (\UnreliableNarrator), containing texts with labeled intra-narrationally, inter-narrationally, and inter-textually unreliable narrators.
We collect short text samples containing first-person narrators from multiple textual domains, including personal accounts from PersonaBank (\textit{Blog post}) \cite{lukin-etal-2016-personabank}, posts from r/AITA\footnote{Posts are scraped between April 2020 and October 2021 from https://www.reddit.com/r/AmItheAsshole/} (\textit{Subreddit}) \cite{vijjini2024socialgaze}, and hotel reviews from Deceptive Opinion\footnote{The Review dataset contains real and fake (deceptive) hotel reviews and is intended for the task of identifying deceptive reviews. Since the Reviews task differs from identifying unreliable narrators, we only collect real reviews for our dataset.} (\textit{Reviews}) \cite{ott2011finding, ott-etal-2013-negative}. We intend to first learn how to classify narrators from a fictional domain and then generalize this knowledge to other textual domains. To this end, we additionally collect about $500$ narrative snippets from stories from Project Gutenberg (\textit{Fiction}).\footnote{https://www.gutenberg.org/}

All text samples are written in first-person (hand-verified) and range from $24$ to $1050$ tokens. Samples from Blog post, Subreddit, and Review contain the entire original written text and are arguably complete narratives. We note that Fiction samples are narrative snippets and do not necessarily contain complete stories with fully developed beginnings, middles, and endings. Table \ref{tab:corpora-stats} shows corpora statistics. Additional details, such as how snippets are selected from the source corpora are given in Appendix \ref{sec:licensing}.

We design an annotation study, determined exempt by the Institutional Review Board, and ask 10 human annotators with undergraduate or graduate degrees in English literature to read and determine the intra-narrational, inter-narrational, and inter-textual unreliabilities of each narrative. We note that this is a time-consuming task: each sample takes annotators roughly 5 minutes each to read, analyze, and annotate. Because the 3 tasks focus on different aspects of the narrative, annotators report having to re-evalutate the narrative for each task. For 817 narratives, each annotated at least twice, we estimate the study took 172 hours. See Appendix \ref{sec:annotation-study-details} for additional details.

Annotators are given the definitions and examples of the three forms of unreliability as described in Section \ref{sec:definitions-of-unreliability}. For each form, they are tasked with choosing the most relevant unreliable label. If none fit, they may decide that the narrator is reliable for that form. For example, for inter-textual unreliability, the annotator is asked to choose one label from ``naïf'', ``madman'', ``pícaro'', ``clown'', or ``none: reliable''. See Appendix \ref{sec:annotator-instructions} for an outline of the instructions given to the annotators. 

Each text sample is annotated by a minimum of two expert annotators. For these pairs of initial results, we calculate inter-annotator agreement with Cohen Kappa's score and observe substantial agreement \cite{landis1977measurement} across all samples: intra-narrational $\kappa=0.75$, inter-narrational $\kappa=0.71$, inter-textual $\kappa=0.73$. We improve label consistency by resolving disagreeing labels: annotators participate in robust conversations\footnote{Annotators either meet via video-call or exchange detailed messages. For disagreeing labels, they discuss their choices and select a final resolved label. If they are unable to agree, a third annotator decides the resolved label, given their arguments. Time spent per discussion: simple texts $\approx$ 2 minutes, samples with very complicated narrators $\approx$ 15-20 minutes.} 
regarding differing labels and choose the best one. Statistics for the distribution of resolved labels are given in Figure~\ref{fig:label-distribution} and additional information, including a numerical breakdown of counts, is given in Appendix \ref{sec:annotated-label-statistics}.

To encourage thoughtful choices, annotators write short descriptions listing observations and brief explanations for why they demonstrate unreliability. All resolved labels have corresponding descriptions; hence, each narrative has three descriptions (1 per unreliability). We calculate across all descriptions an average of $21.2$ tokens, with a maximum of 299 tokens in a given description. See Appendix \ref{sec:unreliability-types-examples} for examples. 

\section{Identifying Unreliable Narrators}

\subsection{Task Definition}

Given $n$, a text narrated by a first-person narrator, we classify narrators for intra-narrational, inter-narrational, and inter-textual unreliability as follows. For intra-narrational unreliability, we want to determine $n\in\{A, R\}$ where $A$ corresponds to $n$ having verbal tics and $R$ corresponds to $n$ not having verbal tics (intra-narrationally reliable). For inter-narrational unreliability, we want to determine $n\in\{A, B, R\}$ where $A$ corresponds to $n$ having a ``same reliable character over time'', $B$ corresponds to $n$ having an ``other character contradiction'', and $R$ corresponds to $n \notin \{A, B\}$ (inter-narrationally reliable). For inter-textual unreliability, we want to determine $n\in\{A, B, C, D, R\}$ where $A$, $B$, $C$, $D$ corresponds to $n$ having a naïf, madman, pícaro, or clown, respectively, and $R$ corresponds to $n \notin \{A, B, C, D\}$ (inter-textually reliable). See Figure~\ref{fig:narrative-example} for an example with the list of classes.

\subsection{Methods}

We seek methods that deal with the complexities of classifying unreliable narrators by learning from snippets from Fiction and testing in an out-of-domain manner on real-world domains. For this purpose, we try zero-shot and few-shot settings, fine-tuning using Parameter-Efficient Fine-Tuning with Low-Rank Adaptation (LoRA) \cite{hu2022lora}, and curriculum learning (CL)
which trains models first on easy and then harder samples.

For CL, the training dataset is divided into easy samples (\textit{Subset-Easy}) and difficult samples (\textit{Subset-Difficult}). We define difficulty of a sample based on how ambiguous it is. Specifically, for each type of unreliability (i.e., intra-narrational, inter-narrational, inter-textual), we observe some samples might contain traits of more than one label. For example, in difficult samples, a narrator who is predominantly a madman might also exhibit some pícaro-like or clown-like traits. Hence, for this sample, in addition to madman, pícaro or clown are also incorrect but reasonable \textit{candidates} for the label.
We hypothesize that samples with fewer candidates are easier to classify because there are fewer potential choices for the final label. Samples with multiple candidates are more challenging because each candidate has an arguable, albeit potentially weak, claim to being chosen as the final label. 

Based on this motivation, we create (\textit{Subset-Easy}) and (\textit{Subset-Difficult}). For this, the LLM is queried to produce a list of counts for the number of traits for each label. For example, for inter-textual unreliability the LLM generates a list such as, \texttt{[A:<NUM>, B:<NUM>, C:<NUM>, D:<NUM>]} where \texttt{A}, \texttt{B}, \texttt{C}, \texttt{D} respectively correspond to naïf, madman, pícaro, and clown, and \texttt{NUM} is the total number of traits present in the narrative for the given label. Candidates are labels with a \texttt{NUM} value $>0$. The training samples are ranked accordingly in order of the least to the most number of candidates. The reordered set is divided in half into \textit{Subset-Easy} and \textit{Subset-Difficult}. An LLM is first fine-tuned on \textit{Subset-Easy} and then on \textit{Subset-Difficult} using LoRA adapters with 8-bit quantization for 3 epochs and default PEFT configuration.

\section{Experiments}

Experiments are performed on Instruct models for Llama3.1-8B, Llama3.3-70B, Mistral-7B, Phi3-medium, GPT-4o mini, and o3-mini (reasoning model). We also compare results with smaller LM classifiers, BERT and ModernBERT. Setup and prompts are described in Appendix~\ref{sec:experimental-setup}. We use Fiction training/validation samples for model training and development and the (remaining) narratives from Fiction, Blog posts, Subreddit, and Reviews as testing samples. In this way, we test on Fiction in an in-domain manner and on the remaining datasets in an out-of-domain manner.

\subsection{Results}
\begin{table*}[t]
    \centering
    \footnotesize
    \begin{tabular}{llcc||ccc}
        \toprule
        && \textbf{CL} & \textbf{Fine-tuned} & \textbf{Zero-Shot} & \textbf{One-Shot} & \textbf{Three-Shot} \\
        \toprule
        \textbf{Intra-nar}  &\hspace{.25cm}\textit{Fiction}     &58.51±1.93	&50.09±1.96	&45.17±1.83	&52.67±2.00	&51.72±2.12\\
                            &\hspace{.25cm}\textit{Blog post}   &53.94±2.22	&50.63±2.27	&45.56±1.80	&29.33±4.48	&40.54±0.73\\ 
                            &\hspace{.25cm}\textit{Subreddit}   &50.04±2.21	&49.00±2.05	&47.41±1.32	&52.03±2.38	&48.87±1.86\\
                            &\hspace{.25cm}\textit{Review}      &67.17±2.16	&55.85±2.35	&58.46±2.29	&60.22±2.20	&52.81±2.25\\
        
        \midrule
        \textbf{Inter-nar}  &\hspace{.25cm}\textit{Fiction}     &34.59±1.82	&34.63±2.26	&16.20±2.19	&15.97±1.19	&17.09±1.26\\
                            &\hspace{.25cm}\textit{Blog post}   &35.92±2.47	&28.73±1.80	&23.15±2.92	&22.19±1.40	&27.46±1.47\\ 
                            &\hspace{.25cm}\textit{Subreddit}   &30.91±1.80	&25.59±1.90	&30.97±1.77	&22.65±1.35	&21.68±1.37\\
                            &\hspace{.25cm}\textit{Review}      &35.29±1.66	&36.59±2.18	&25.85±1.79	&25.67±3.11	&25.37±3.10\\
                            
        \midrule
        \textbf{Inter-tex}  &\hspace{.25cm}\textit{Fiction}     &27.42±1.87	&28.59±1.87	&18.22±2.38	&24.00±1.55	&23.54±1.69\\
                            &\hspace{.25cm}\textit{Blog post}   &19.58±1.78	&18.99±1.34	&24.23±2.79	&28.59±1.75	&24.35±1.56\\ 
                            &\hspace{.25cm}\textit{Subreddit}   &13.49±1.55	&10.85±1.31	&12.95±1.21	&12.01±1.11	&10.71±1.14\\
                            &\hspace{.25cm}\textit{Review}      &16.72±0.67	&17.54±1.35	&15.75±1.31	&20.32±1.08	&19.30±2.08\\
        \bottomrule
        
    \end{tabular}
    \caption{Breakdown of unreliability F1 (macro) scores for each domain for Llama3.1-8B. Improvements on left are statistically significant compared to results on right row-wise with $p<0.05$ \cite{dror2018hitchhiker}.}
    \label{tab:f1scores-breakdown}
\end{table*}
\begin{table*}[t]
    \centering
    \footnotesize
    \begin{tabular}{llcc||ccc}
        \toprule
        && \textbf{CL} & \textbf{Fine-tuned} & \textbf{Zero-Shot} & \textbf{One-Shot} & \textbf{Three-shot} \\
        \toprule
        \textbf{Intra-nar}  &\textit{Llama3.1-8B}     &57.42±2.13	&51.39±2.16	&49.15±1.81	&48.56±2.76	&48.48±1.74 \\
                            &\textit{Llama3.3-70B}    &51.26±2.12	&51.28±2.09	&54.20±1.65	&63.89±2.28	&61.41±1.91 \\
                            &\textit{Mistral-7B}      &55.76±1.70	&56.46±2.11	&56.79±2.05	&50.87±1.96	&52.99±2.24 \\
                            &\textit{Phi3-medium}    &53.75±2.14	&52.18±2.36	&60.00±2.22	&44.70±1.69	&44.86±1.49 \\
                            &\textit{GPT-4o mini}     &---    &---	&47.88±2.05	&50.51±1.67	&51.77±2.25 \\ 
                            &\textit{o3-mini}         &---    &---	&42.22±1.97	&43.47±2.00	&44.32±2.04 \\
        
        \midrule
        \textbf{Inter-nar}  &\textit{Llama3.1-8B}     &34.18±1.94	&31.39±2.03	&24.04±2.17	&21.62±1.76	&22.90±1.80 \\
                            &\textit{Llama3.3-70B}    &33.49±2.31	&30.32±1.29	&29.11±1.63	&31.23±1.72	&34.02±2.23 \\
                            &\textit{Mistral-7B}     &31.15±1.45	&25.75±0.44	&19.49±1.36	&33.07±1.92	&31.29±1.86 \\
                            &\textit{Phi3-medium}    &22.32±1.49	&35.76±1.81	&25.23±1.88	&23.42±1.71	&24.66±1.73 \\
                            &\textit{GPT-4o mini}     &---    &---	&28.15±1.49	&31.48±1.70	&26.00±1.52 \\
                            &\textit{o3-mini}         &---    &---	&32.18±1.90	&28.79±0.91	&27.40±1.59 \\
                            
        \midrule
        \textbf{Inter-tex}&\textit{Llama3.1-8B}     &19.30±1.47	&18.99±1.47	&17.79±1.92	&21.23±1.37	&19.48±1.62 \\
                            &\textit{Llama3.3-70B}    &21.04±1.69	&21.02±1.64	&28.52±1.96	&30.80±1.81	&28.23±1.89 \\
                            &\textit{Mistral-7B}      &29.68±2.01	&24.38±1.29	&20.23±1.51	&18.35±1.43	&17.12±1.35 \\
                            &\textit{Phi3-medium}    &25.00±1.51	&26.24±1.82	&27.56±1.70	&18.84±1.84	&16.41±1.38 \\
                            &\textit{GPT-4o mini}     &---    &---	&17.84±1.41	&20.66±1.42	&19.98±1.38 \\
                            &\textit{o3-mini}         &---    &---	&16.65±1.14	&15.44±0.33	&15.84±1.54 \\
        
        \bottomrule
    \end{tabular}
    \caption{Unreliability F1 (macro) scores for combined domains for all model families and sizes. Results on left are statistically significant compared to results on right row-wise.}
    \label{tab:f1scores}
\end{table*}

\begin{table}[t]
    \footnotesize
    \centering
    \begin{tabular}{llcc}
        \toprule
        && \textbf{BERT} & \textbf{ModernBERT} \\
        \toprule
        \textbf{Intra-nar}  &\textit{Fiction} & 48.42 & 49.48\\
                            &\textit{Avg} & 17.77 & 39.94\\
        
        \midrule
        \textbf{Inter-nar}  &\textit{Fiction} & 31.37 & 38.46\\
                            &\textit{Avg} & 25.76 & 27.07\\
                            
        \midrule
        \textbf{Inter-tex}  &\textit{Fiction} & 12.46 & 14.71\\
                            &\textit{Avg} & 11.12 & 16.98\\
        \bottomrule
        
    \end{tabular}
    \caption{Unreliability F1 (macro) scores for Fiction and combined domains for smaller LM classifiers.}
    \label{tab:lm-results}
\end{table}

Table \ref{tab:f1scores-breakdown} presents performances of CL, fine-tuned, zero-shot, and few-shot methods where macro-averaged F1 scores are provided for each domain (using Llama3.1-8B). Table \ref{tab:f1scores} presents the performance of LLMs averaged across domains, and Table \ref{tab:lm-results} shows the performance of LM classifiers.

We notice six key takeaways. First, generally speaking, all methods and models perform better for the intra-narrational task than for the other two tasks. Similarly, they perform better for the inter-narrational task than for inter-textual. This finding indicates the intra-narrational task is easiest, and the inter-textual task (requiring more abstract inferences) is most difficult for LLMs. Appendix~\ref{sec:demonstration-of-task-complexities} shows an example demonstrating how the inter-textual task requires a deeper understanding of the narrator's state of mind, making it more difficult.

Second, methods using training samples (i.e., CL, fine-tuning, few-shot) outperform the zero-shot method, indicating that training data does improve LLM performance. Appendix \ref{sec:effect-of-learning-from-examples} shows samples where incorrect labels are predicted in zero-shot and correct labels are predicted in few-shot because the model learns from the shots. 

Third, for most cases, CL outperforms fine-tuning, indicating that more sophisticated ways of leveraging the training data is promising for better performance. 

Fourth, in Table \ref{tab:f1scores-breakdown}, we observe that out-of-domain performances, especially those whose methods use more training data (i.e., CL and fine-tuning), are not better but good compared to in-domain performances. This result indicates that it is possible to learn from the Fiction text domain and apply that knowledge to other real-world text domains. We make similar observations for other models (not shown here due to space constraints). 

Fifth, Table \ref{tab:f1scores} shows CL improves performance of smaller models but not larger ones. E.g., Llama3.3-70B few-shot performs competitively with CL and fine-tuning, indicating that as model size increases, learning from fewer samples yields comparable predictive capabilities to learning with more samples.

Finally, for experiments with LM classifiers, we observe average values across all test sets are less than average values for CL and fine-tuned methods. These results indicate that LMs do not outperform LLMs. To understand these observations, we note that for Fiction (in-domain experiment), LMs give results comparable to our zero-shot method; however, for all the other test sets (out-of-domain experiment), the performance of the LMs drastically drops. Hence, we determine that LMs are less capable than LLMs of generalizing knowledge learned from one domain to other domains.

We provide individual performance breakdowns of the remaining LLMs for each domain in Appendix \ref{sec:all-model-results} (including a breakdown of class-wise scores in Table \ref{tab:class-wise}) and an error analysis of incorrectly classified narrators in Appendix \ref{sec:errors-classifying-un}.

\section{Analysis}
\label{sec:analysis}
In this section we analyze unreliability classification with respect to various narrative properties: narrator's gender, number of characters, narration style, and overall narrative sentiment. For these experiments, we use CL outputs for one model from each open-source LLM family ($3$ total models). We use Llama3.3-70B to automatically infer these narrative properties (the complete prompts and an error analysis are given in Appendix \ref{sec:analysis-prompts} and \ref{sec:prompting-for-analysis}).\\

\begin{figure}[t]
    \centering
    \includegraphics[width=\linewidth, height=170pt]{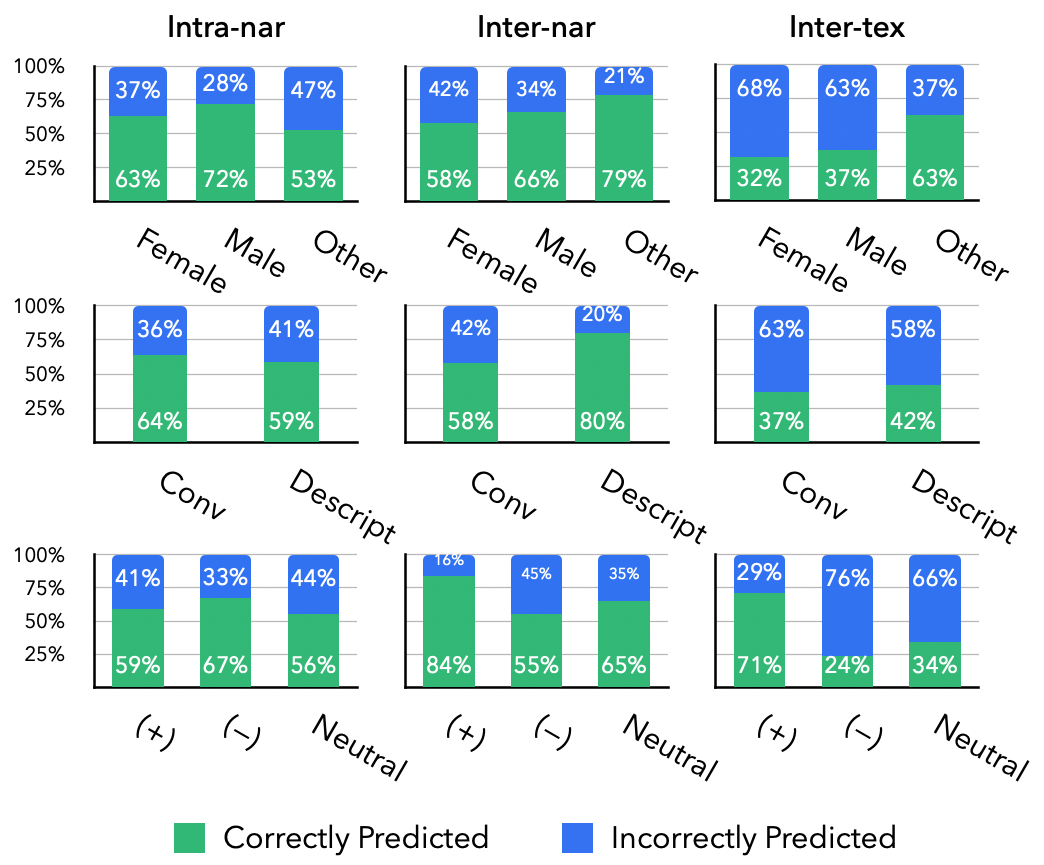}
    \caption{Breakdown of correctly predicted (green) vs. incorrectly predicted (blue) unreliable narrators. Top row: with respect to the narrator's gender $\in$ \{female, male, other\}. Middle row: with respect to narrative style $\in$ \{conversational, descriptive\}.  Bottom row:  with respect to narrative sentiment tone $\in$ \{positive, negative, neutral\}. Results are from Llama3.1-8B experiments.}
    \label{fig:analysis}
\end{figure}

\noindent\textbf{\textit{RQ1: Does the gender of the narrator affect the prediction?}} Across all testing samples, we count 125 female, 215 male, and 43 other/ambiguous narrators. The first row of Figure~\ref{fig:analysis} shows the percentages of female, male, and other narrators classified w.r.t. unreliable narrators correctly vs. incorrectly by Llama3.1-8B for 5 runs across all testing samples. Figure~\ref{fig:analysis-gender} in Appendix \ref{sec:additional-analysis-results} shows results from other models. We observe across all model families that male narrators are predicted correctly more frequently than female narrators. For inter-narrational and inter-textual tasks, other/ambiguous characters are predicted more correctly than either female or male narrators, indicating that performance improves when the narrator is not specified as female or male. \\

\begin{figure}[t]
    \centering
    \includegraphics[height=130pt]{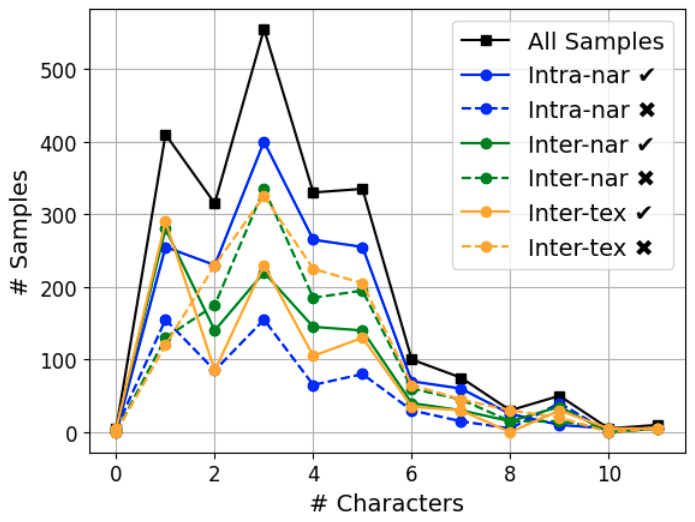}
    \caption{Number of characters vs. number of samples. All Samples (solid black) is the distribution of all narratives with respect to the number of characters. Blue, green, and orange solid lines show correct predictions, and corresponding dashed lines show incorrect predictions. Results are from Mistral experiments.}
    \label{fig:analysis-num-char-mistral}
\end{figure}

\noindent\textbf{\textit{RQ2: How does the narration style change the difficulty of the prediction?}} The middle row of Figure~\ref{fig:analysis} shows that narratives written in a conversational style tend to perform slightly better than those written in a descriptive style for intra-narrational unreliability. This could be because it might be easier to detect the verbal tics within a conversational tone. However, for inter-narrational and inter-textual tasks, narratives written in a descriptive tone perform better. Figure~\ref{fig:analysis-gender} in Appendix \ref{sec:additional-analysis-results} shows results from other models.\\

\noindent\textbf{\textit{RQ3: How does the overall narration sentiment affect the prediction?}} The last row of Figure~\ref{fig:analysis} demonstrates that narratives written in a negative tone perform better than narratives written in a positive tone for intra-narrational unreliability. This result is likely a consequence of negative tones often harboring multiple verbal tics, resulting in an easier prediction. For inter-narrational and inter-textual unreliabilities, narratives written in a positive tone result in significantly better predictions than narratives written in a negative tone. See Figure~\ref{fig:analysis-tone} in Appendix \ref{sec:additional-analysis-results} for results from other models.\\

\noindent\textbf{\textit{RQ4: Are narratives with multiple characters trickier to predict?}} Figure~\ref{fig:analysis-num-char-mistral} shows the majority of narratives contain 1-5 characters. Within this range, correct predictions for unreliabilities (solid blue, green, yellow) peak at narratives with 1 and 3 characters. For intra-narrational classification, there are consistently more correct (solid blue) than incorrect (dotted blue) predictions, indicating the number of characters does not change the difficulty of the narratives to classify. For inter-narrational and inter-textual classification, the number of incorrectly predicted narratives (dotted green and yellow) surpasses the number of correctly predicted narratives (solid green and yellow) when the number of characters $\ge2$, suggesting that narratives with multiple characters are tricker to predict than narratives with only the narrator. See Figure~\ref{fig:analysis-num-char-llama} and Figure~\ref{fig:analysis-num-char-phi} in Appendix \ref{sec:additional-analysis-results} for other model results.

Additional details regarding our methods of performing analysis are given in Appendix \ref{sec:details_of_analysis}.

\section{Conclusion}
We propose using automatic methods to classify intra-narrationally, inter-narrationally, and inter-textually unreliable narrators. Borrowing definitions from narratology we define binary and multi-class classification tasks, annotate narratives from a diverse domain of texts, and evaluate the ability of LLMs to perform these classification tasks in zero-shot, few-shot, fine-tuned, and curriculum learning settings. We observe that these tasks are very tricky for LLMs to solve and offer our findings as a call for future work to further investigate the use of NLP methods to identify unreliable narrators.

\section{Limitations}

Firstly, this work focuses on short texts (no longer than 1050 tokens each), some of which do not contain complete beginnings, middles, and endings. We encourage future work to consider this task for longer-length texts, such as full-length short stories or books. Secondly, we note that all samples in our datasets are written in English. As the definitions of unreliability are applicable to works of other languages, we recommend future work exploring this task on other languages. Thirdly, for \textit{RQ1} in Section~\ref{sec:analysis}, we limit our analysis to only female, male, and other/ambiguous genders. Finally, we observe that the size of the dataset is relatively small due to the high cost of high-quality annotations.

\section*{Acknowledgments}
We are grateful for the suggestions from our anonymous reviewers, and we thank Haoyuan Li, Anvesh Rao Vijjini, Somnath Basu Roy Chowdhury, and Amartya Banerjee for their discussions and valuable insights. This work was supported in part by NSF grant IIS2047232.

\bibliography{acl_latex}

\appendix

\section{Examples of Unreliable Narrator Types}
\label{sec:unreliability-types-examples}
Samples from the training set are given in Sections~\ref{sec:intra-nar-training-sample}, \ref{sec:inter-nar-training-sample}, \ref{sec:inter-tex-training-sample}. Samples from the testings sets are given in Sections~\ref{sec:intra-nar-test-sample}, \ref{sec:inter-nar-test-sample}, \ref{sec:inter-tex-test-sample}. 

\subsection{Intra-narrational Training Sample}
\label{sec:intra-nar-training-sample}

\noindent\underline{\textbf{Verbal Tics Present:}} \textit{(Fiction)} ``Don't you 'mais' me, sir! I had two trunks—deux troncs—when I got aboard that wabbly old boat at Dover this morning, and I'm not going to budge from this wharf until I find the other one. Where did you learn your French, anyway? Can't you understand when I speak your language?'' \\

\noindent\textit{Explanation:} Defensive tone throughout. Digression: ``Where did you learn your French anyway?''

\subsection{Inter-narrational Training Samples}
\label{sec:inter-nar-training-sample}

\noindent\underline{\textbf{Same Unreliable Narrator Over Time:}} \textit{(Fiction)}
``This is written from memory, unfortunately. If I could have brought with me the material I so carefully prepared, this would be a very different story. Whole books full of notes, carefully copied records, firsthand descriptions, and the pictures—that’s the worst loss. We had some bird’s-eyes of the cities and parks; a lot of lovely views of streets, of buildings, outside and in, and some of those gorgeous gardens, and, most important of all, of the women themselves.''\\

\noindent\textit{Explanation:} The narrator is reflecting back on events in the past where narrator has admitted unreliability. There is no indication of change in reliability over time.\\

\noindent\underline{\textbf{Other Character Contradiction:}} \textit{(Fiction)}
``My kinsman and myself were returning to Calcutta from our Puja trip when we met the man in a train. From his dress and bearing we took him at first for an up-country Mahomedan, but we were puzzled as we heard him talk. He discoursed upon all subjects so confidently that you might think the Disposer of All Things consulted him at all times in all that He did. Hitherto we had been perfectly happy, as we did not know that secret and unheard-of forces were at work, that the Russians had advanced close to us, that the English had deep and secret policies, that confusion among the native chiefs had come to a head. But our newly-acquired friend said with a sly smile: ``There happen more things in heaven and earth, Horatio, than are reported in your newspapers.'' As we had never stirred out of our homes before, the demeanour of the man struck us dumb with wonder. Be the topic ever so trivial, he would quote science, or comment on the Vedas, or repeat quatrains from some Persian poet; and as we had no pretence to a knowledge of science or the Vedas or Persian, our admiration for him went on increasing, and my kinsman, a theosophist, was firmly convinced that our fellow-passenger must have been supernaturally inspired by some strange ``magnetism'' or ``occult power,'' by an ``astral body'' or something of that kind. He listened to the tritest saying that fell from the lips of our extraordinary companion with devotional rapture, and secretly took down notes of his conversation. I fancy that the extraordinary man saw this, and was a little pleased with it.''\\

\noindent\textit{Explanation:} The new friend contradicts what the narrator believes is occurring in the world.

\subsection{Inter-textual Training Samples}
\label{sec:inter-tex-training-sample}

\noindent\underline{\textbf{Naïf:}} \textit{(Fiction)}
``They went off and I got aboard the raft, feeling bad and low, because I knowed very well I had done wrong, and I see it warn’t no use for me to try to learn to do right; a body that don’t get started right when he’s little ain’t got no show—when the pinch comes there ain’t nothing to back him up and keep him to his work, and so he gets beat. Then I thought a minute, and says to myself, hold on; s’pose you’d a done right and give Jim up, would you felt better than what you do now? No, says I, I’d feel bad—I’d feel just the same way I do now. Well, then, says I, what’s the use you learning to do right when it’s troublesome to do right and ain’t no trouble to do wrong, and the wages is just the same? I was stuck. I couldn’t answer that. So I reckoned I wouldn’t bother no more about it, but after this always do whichever come handiest at the time.'' \\

\noindent\textit{Explanation:} Narrator appears to have made a mistake and has become a foil to a lamentable social condition. They clearly do not understand the complexity of their environment: ``Well, then, says I, what’s the use you learning to do right when it’s troublesome to do right and ain’t no trouble to do wrong, and the wages is just the same? I was stuck. I couldn’t answer that.''\\

\noindent\underline{\textbf{Madman:}} \textit{(Fiction)}
``You lie, cursed dog! What a scandalous tongue! As if I did not know that it is envy which prompts you, and that here there is treachery at work—yes, the treachery of the chief clerk. This man hates me implacably; he has plotted against me, he is always seeking to injure me. I'll look through one more letter; perhaps it will make the matter clearer.''\\

\noindent\textit{Explanation:}
Narrator uses a frantic voice with accusations and exclamation points and seems maddened by perceived torture (the second character has said things which the narrator does not like). The narrator has strong negative feelings towards others and appears to feel alienated from the rest of the characters, including the chief clerk.\\

\noindent\underline{\textbf{Pícaro:}} \textit{(Fiction)}
``I grew the greatest artist of my time and worked myself out of every danger with such dexterity, that when several more of my comrades ran themselves into Newgate presently, and by that time they had been half a year at the trade, I had now practised upwards of five years, and the people at Newgate did not so much as know me; they had heard much of me indeed, and often expected me there, but I always got off, though many times in the extremest danger.'' \\

\noindent\textit{Explanation:} Narrator fits picaro with rise/fall of fortune and roguish characteristics. Seems to approach problems cleverly and with clear motivations.\\

\noindent\underline{\textbf{Clown:}} \textit{(Fiction)}
``Therefore, my dear friend and companion, if you should think me somewhat sparing of my narrative on my first setting out—bear with me,—and let me go on, and tell my story my own way:—Or, if I should seem now and then to trifle upon the road,—or should sometimes put on a fool’s cap with a bell to it, for a moment or two as we pass along,—don’t fly off,—but rather courteously give me credit for a little more wisdom than appears upon my outside;—and as we jog on, either laugh with me, or at me, or in short do any thing,—only keep your temper.''\\

\noindent\textit{Explanation:} Narrator flips the narrative of his past (``if you should think me somewhat sparing of my narrative on my first setting out—bear with me,—and let me go on, and tell my story my own way'') and seems intent to repackage conflict in a new light.

\subsection{Intra-narrational Testing Sample}
\label{sec:intra-nar-test-sample}
\noindent\underline{\textbf{Verbal Tics Present:}} \textit{(Subreddit)} 
``The guy I bought the truck from patched it together for the sale at first small stuff like not vacuuming the AC system or armrest falling off the door from double sided tape, but then one day the throattle got stuck wide open I pulled it out of gear killed the ignition pulled over rip apart the throattle body open and found a plastic cap crammed in there... I text the owner about it and he said ``wrong number'' I live in a metropolitan area and have been putting off transferring title and racking up tolls at first I was vengefully self righteous but now I'm over thinking it. Am I the asshole?''\\

\noindent\textit{Explanation:} Example of admission of fault/bias: “I live in a metropolitan area and have been putting off transferring title and racking up tolls at first I was vengefully self righteous but now I'm over thinking it. Am I the asshole?”

\subsection{Inter-narrational Testing Samples}
\label{sec:inter-nar-test-sample}
\underline{\textbf{Same Unreliable Narrator Over Time:}} \textit{(Blog post)}
``I got a haircut yesterday and I hate hate HATE it! I kind of got it done on impulse (i mean I needed it done anyway) so I wasn't prepared with photos or anything and I had to flip through a magazine looking for what i wanted. I wanted a short version of this basically. Like where the top layer curled into my chin. I found similar enough haircuts in a magazine and told her what I wanted. It looked fine wet but it dried completely wrong. She cut it way too short in the front and its basically a shorter version of what I just had WHICH IS NOT WHAT I WANTED. Because my hair is f****** THICK so short hair is f****** POOFY. And I dunno I just think I look really stupid now. It's just completely wrong and I had her layer the back, which by the way, she didn't even do that right. Just uuuuuugh I should have just gotten my bangs trimmed or something and idk I want to cry now. And I'm not even depressed about that. Just being on campus for a week has made me feel so f****** shitty. Like shittier than I felt shut up in my room all summer. ps I realized I'm scared to wear lolita on campus now. awesome y/y :|''\\

\noindent\textit{Explanation:} Narrator is recounting event from the past and demonstrates intra-narrational unreliability without indicating growth or change over time.\\

\noindent\underline{\textbf{Other Character Contradiction:}} \textit{(Review)}
``I stayed at the Monaco-Chicago back in April. I was in town on business, and the hotel was recommended by a friend of mine. Having spent a weekend there, I have no idea what my friend was talking about. The complimentary morning coffee was weak; the fitness room was dimly lit; and I thought I'd have to have my clothes mailed back to me when I used their supposed 'overnight' laundry service for a suit I spilled some wine on. My room was adequate, but nowhere near what I've seen elsewhere at this price point. Recent renovation must be slang for 'everything is stiff and smells of industrial adhesive.' The mattress in my room was incredibly firm, and I slept poorly. When I travel, I expect an experience similar to or better than my experience at home. At most hotels, I receive excellent service and comfortable accomodations. This was an exception to my usual, and I won't be back anytime soon.''\\

\noindent\textit{Explanation:} The friend who recommended the hotel is in contradiction with the narrator.\\

\subsection{Inter-textual Testing Samples}
\label{sec:inter-tex-test-sample}

\noindent\underline{\textbf{Naïf:}} \textit{(Subreddit)}
``My parents are 80 years old. Lately they both have been having some health issues and it appears they are finalizing/updating their Last Will \& Testaments. I, (44f) am the youngest of three girls. Earlier this year my mother came up for a visit and she was staying at my house. During her first night she told me some of the health issues my father has been experiencing. Later that night she told me she had three items that were special to her (her wedding ring, a painting, and some china my father bought in Eqypt when he was in the service). I don't know if its relevant but they are all worth roughly the same value, a modest sum, nothing over the top. She asked me which one I would like to inherit when the time came. The whole conversation was awkward and uncomfortable but in the end I selected the wedding ring.
  
A few months later, my two older sisters and I were out at a bar and this topic came up. Apparently they were angry with me about choosing the ring. They say ``Who does that? Why did you even pick one?''. I was taken aback because I didn't even know this was an issue with them but apparently they talked about it amongst themselves. I told them I don't understand why they were so angry with me. Mom asked me which I would like so I told her if given the option I would like the ring. 
  
We all have daughters so i understand why my sisters might also want the ring. However, how can I be held accountable for the fact that my mom gave me the first choice and so I choose?
  
AITAH for choosing to inherit the ring when my mother specifically asked me which item I would prefer?’’ \\

\noindent\textit{Explanation:} The narrator's experience exposes ``a lamentable social condition'' (she is dealing with aging, possibly even dying, parents and is put in the ``awkward and uncomfortable'' situation of choosing her inheritance in the midst of bickering siblings). Does our narrator lack the ``experience to fully understand the narrated events or the complexity of their environment'' per our revised naif definition? One could say yes. That she is surprised by all this drama suggests this is the first time she is in this situation and so she is inexperienced in this matter. Also, she is struggling with understanding the complexity of her environment (arguing siblings): ``I was taken aback because I didn't even know this was an issue with them but apparently they talked about it amongst themselves. I told them I don't understand why they were so angry with me. Mom asked me which I would like so I told her if given the option I would like the ring.... how can I be held accountable for the fact that my mom gave me the first choice and so I choose?''\\

\noindent\underline{\textbf{Madman:}} \textit{(Blog post)}
``Today I realized how much I actually love her, and no not that crazy chick that I finally got rid of. I was talking on the phone with her and we had run out of things to talk about so I had started singing Lips of an Angel. She'd occasionally sing along but I don't think she was paying much attention to it. But the whole damn song I was picturing her. The lyrics just kept screaming out at me. I mean, not how the song was intended. Its really about two people breaking up and the girl calling up her ex and they're talking about how they still have feelings for each other. But the way I took it in was how her boyfriend doesn't like us talking but she kept talking to me anyways and all these stupid feelings I have for her. I don't even know. I'm just an idiot, I guess. But it gotten to the point I started crying. I didn't let her know that because I already made things awkward enough in the past and I didn't want to bring it up again. Because she only looks at me as a friend and as much as it kills me...thats all we'll ever be.''\\

\noindent\textit{Explanation:} Narrator is a madman who demonstrates high emotions. He seems maddened by the perceived torture of not being able to date a girl he thinks he is in love with. He has really strong feelings that can affect his sanity (“I already made things awkward enough in the past”).\\

\noindent\underline{\textbf{Pícaro:}} \textit{(Review)}
``I want to issue a travel-warning to folks who might sign up for the weekend deal they offer through travelzoo from time to time: The deal says 'free breakfast' included in the price. However, what they don't tell you, is that the breakfast consists of a cup of coffee and a bisquit (or two)! Moreover, you need to ask for these 'tickets' at the lobby when you check in - they won't give them to you automatically! We stayed there over Christmas '03, and we, and I noticed several guests who bought the same package, had a rather unpleasant experience! The hotel is nice though, if you don't consider their lousy service!''\\

\noindent\textit{Explanation:} The clever vibes stick out to me a little bit. Rather than just stating that ``travelzoo'' is lame, the narrator writes ``I want to issue a travel-warning.'' Also, the air quotes around free breakfast and tickets also read to me as clever. And the exclamation, parenthetical, and slyness in the following sentence also seems to suggest cleverness: ``what they don't tell you, is that the breakfast consists of a cup of coffee and a bisquit (or two)!'' The narrator also states in an understated way that them and other guests who bought this package ``had a rather unpleasant experience!'' with ``rather'' suggesting trying to be clever too. And lastly, ``The hotel is nice though, if you don't consider their lousy service!'' feels like the last biting remark attempting cleverness.\\

\noindent\underline{\textbf{Clown:}} \textit{(Subreddit)}
``Some guy in class, who I'm sort-of close to, had planned his birthday and wanted to celebrate it with all of us in the classroom.
 
He brought us all goodie-bags, and even brought a cake. At first, I was annoyed. Because I thought this was kind of excessive for a birthday but I kept quiet because I didn't want to hurt his feelings. And since I disliked accepting gifts from others, I refused to take the goodie-bag.
  
He had talked about it with the teacher and let all of us take slices of his cake and passed it around to everyone. I personally disliked cake so I said no, along with some few 2 or 3 other people.
  
He understood where we were coming from so he didn't take it to heart. After everyone had enjoyed their cake, the teacher had asked us all to group up to take a picture. Everyone started huddling in the middle but since I didn't feel comfortable taking a picture with a bunch of people, I just kept sitting on my seat and politely refused to take a picture with them.
  
They just accepted it, it was all good until my parents scolded me for not taking a picture with them (since the picture was posted on Instagram) and she didn't want the other parents to think I was rude or inconsiderate. She told me to apologize the next time I saw him and I agreed.
 
Time-skip to the day after, I went up to him and apologized if he thought I didn't like him and told him that I felt uncomfortable with crowds of people, as well as accepting things that weren't mine. He didn't mind and said it was alright, that he understood how I felt but he said it in a really sad voice that broke my heart. AITA?’’\\

\noindent\textit{Explanation:} Narrator is a clown who re-interprets the situation. Instead of considering the events as a time of celebration, the narrator focuses on his/her personal viewpoints (e.g., refusing to take the goodie bag because narrator does not like to accept gifts, not eating cake because narrator personally dislikes it, and not joining the class photo because narrator didn’t feel comfortable in crowds).

\section{Annotation Study Details}
\label{sec:annotation-study-details}

We hire 10 expert workers who are in the process or have completed a university degree (honors bachelor's, master's, PhD) in English literature and have had previous experience analyzing narrators. Workers are paid $\$7.25$/hour (equivalent to the state minimum wage). They provide written consent and understanding of the task before beginning and may complete as many narratives as they are able within a set number of hours. They are free to stop at any point of the process. During the study, no personal information is collected from the annotators.

\subsection{Instructions for Annotators}
\label{sec:annotator-instructions}
Each annotator is given a set of narratives and a list of definitions for intra-narrational, inter-narrational, and inter-textual unreliabilities (same as definitions and examples given in Section~\ref{sec:definitions-of-unreliability}). They agree not to use AI assistants while choosing labels. To ensure consistency across labeling, annotators are instructed:

\textit{``Classify narrators according to the given definitions. Focus on characteristics of the narrator, not on the situation described. Classify using defining characteristics within the narrative only. Do not create hypotheticals to fill in missing details. Assume the narrator is reliable unless unreliable traits are present in the narrative.''}\\

Annotators are tasked with labeling each narrative in the following steps:

\begin{enumerate}
    \item In Column C, read the narrative.
    \item In Column D, list any observed verbal tics (this is a space for notes, or entering the corresponding letter):
    \begin{enumerate}
        \item Admission of fault/bias
        \item Defensive tone
        \item Digressions
        \item Hedging Language
        \item Inconsistencies
        \item Selective Memory
        \item Statement of Potential Disbelief
    \end{enumerate}
    \item In Column E, select the best option for intra-narrational unreliabilities (choose the corresponding letter):
    \begin{enumerate}
        \item Verbal tic(s) present
        \item None: intra-narrationally reliable
    \end{enumerate}
    \item In Column F, select the best option for inter-narrational unreliabilities (choose the corresponding letter):
    \begin{enumerate}
        \item Same unreliable character over time
        \item Other character contradiction
        \item None: inter-narrationally reliable
    \end{enumerate}
    \item In Column G, select the best option for inter-textual unreliabilities (choose the corresponding letter):
    \begin{enumerate}
        \item Naïf
        \item Madman
        \item Pícaro
        \item Clown
        \item None: inter-textually reliable
    \end{enumerate}
    \item In Column H, please leave annotator's notes. This is a space for any additional comments you might have. Feel free to use it or leave it blank.
\end{enumerate}

\section{Annotated Label Statistics}
\label{sec:annotated-label-statistics}

We report the distribution of labels chosen by the human annotators for each text domain for intra-narrational, inter-narrational, and inter-textual unreliabilities in Table \ref{tab:intra-narrational-label-stats}.

\begin{table*}[t]
  \centering
  \footnotesize
  \begin{tabular}{l|cc|ccc|ccccc}

    & \multicolumn{2}{c|}{\textbf{Intra-nar}} & \multicolumn{3}{c|}{\textbf{Inter-nar}} & \multicolumn{5}{c}{\textbf{Inter-tex}}\\
    \toprule
    \textbf{Corpus} & \textbf{(A)} & \textbf{(R)} & \textbf{(A)} & \textbf{(B)} & \textbf{(R)} & \textbf{(A)} & \textbf{(B)}& \textbf{(C)} & \textbf{(D)} & \textbf{(R)}\\
    \toprule
    Fiction     & 264 & 235     &   48&38&413   &   49&76&64&75&235\\
    \hspace{.25cm}\textit{Train/Valid}  & 180 & 193     &   41&31&301   &   37&59&49&50&178\\
    \hspace{.25cm}\textit{Test}         & 84 & 42       &   7&7&112     &   12&17&15&25&57\\
    Blog posts  & 75 & 31       &   36&6&64     &   11&23&10&19&43\\
    Subreddit   & 97 & 15       &   18&63&31    &   29&28&15&24&16\\
    Reviews     & 52 & 48       &   2&5&93      &   3&24&7&3&63\\
    \bottomrule
  \end{tabular}
  \caption{Label statistics (numerical breakdown) for each type of unreliability across all text domains. For \textit{intra-narrational }unreliabilities: Total number of narratives with \textbf{(A)} verbal tics or \textbf{(R)} none (intra-narrationally reliable). For \textit{inter-narrational} unreliabilities: Total number of narratives with \textbf{(A)} ``Same unreliable character over time'', \textbf{(B)} ``Other character contradiction'', or \textbf{(R)} none (inter-narrationally reliable). For \textit{inter-textual} unreliabilities: Total number of narratives with \textbf{(A)} Naïf, \textbf{(B)} Madman, \textbf{(C)} Pícaro, \textbf{(D)} Clown, or \textbf{(R)} none (inter-textually reliable).}
  \label{tab:intra-narrational-label-stats}
\end{table*}

\section{Experimental Setup}
\label{sec:experimental-setup}

We describe licenses of source corpora in Section~\ref{sec:licensing} and implementation in Section~\ref{sec:implementation-details}. All prompts are provided in Section~\ref{sec:prompts-for-experiments}.

\subsection{Source Corpora Details}
\label{sec:licensing}
\begin{table}[t]
    \footnotesize
    \centering
    \begin{tabular}{ll}
        \textbf{Corpora} & \textbf{License} \\
        \toprule
        Project Gutenberg   & Project Gutenberg License\\
        \midrule
        Personabank         &  Creative Commons Attribution\\
                            & 4.0 International License\\
        \midrule
        AITA \cite{vijjini2024socialgaze} & Creative Commons Attribution\\
                            & 4.0 International License\\
        \midrule
        Deceptive Opinion   & Creative Commons Attribution\\
                            & -NonCommercial-ShareAlike\\
                            & 3.0 Unported License\\
        \bottomrule
    \end{tabular}
    \caption{Licenses for the sources of each textual domain in \UnreliableNarrator.}
    \label{tab:corpora_details}
\end{table}

We report the license of each source of narratives for \UnreliableNarrator\ in Table \ref{tab:corpora_details}. All data is used for research purposes only, consistent with their intended use. Narratives were checked to ensure there is no personally identifying information or offensive content. All samples are in English.

We select snippets from the source corpora in the following ways. For train-test splits, snippets are selected at random because we desire class balances that are representative of each domain. We do not use LLMs to pre-select snippets because this process would most likely favor snippets that are easier to classify and would not choose trickier snippets. For selecting demonstration examples in few-shot methods, we acknowledge that a more careful shot selection could potentially help the LLM make better predictions. However, this selection process requires careful research that analyzes narratives from the perspective of various elements like characters, setting, etc. While this would be a good direction for future work, it is out of the scope of the current work.

\subsection{Implementation Details}
\label{sec:implementation-details}

Our experiments are conducted using up to four 48GB Nvidia RTX A6000 GPUs and 2 Nvidia A100-80G GPUs. Open-source checkpoints for Llama, Mistral, Phi3, BERT, and ModernBERT models are obtained from HuggingFace \cite{wolf2019huggingface} library. API-based GPT models are obtained from OpenAI API. See Table~\ref{tab:model_checkpoints} for checkpoint versions and model sizes. For all models, we use a temperature value of 0.7 and top-p value of 0.9 (chosen after experimentation with higher and lower values). The model o3-mini \texttt{reasoning\_level} is left at the default \textit{medium} level. Inference time for a single experiment on an open source model takes approximately 30 minutes. Fine-tuning a LoRA \cite{hu2022lora} adapter for 3 epochs takes approximately 1 hour. Training BERT and ModernBERT models for 3 epochs takes approximately 1 minute.

In addition to utilizing HuggingFace, PEFT, PyTorch libraries for model inference and fine-tuning, we utilize existing Python packages such as PyTorch, pandas, re, scikit-learn, and statistics for pre/post-processing of data and analysis of results.

F1 (macro) scores for all methods are calculated using a bootstrapping evaluation method. For a testing set containing $m$ samples, we perform inference on all $m$ samples 5 times to predict the intra-narrational, inter-narrational, and inter-textual labels. From these 5 runs, 1000 output predictions are randomly sampled with replacement and compared to the corresponding gold labels.

\begin{table}[t]
    \footnotesize
    \centering
    \begin{tabular}{lll}
        \textbf{Alias}  & \textbf{Model Name} & Size\\
        \toprule
        Llama3.1-8B   & meta-llama/ \\&Meta-Llama-3.1-8B-Instruct & 8B\\
        \midrule
        Llama3.3-70B   &meta-llama/ \\&Llama-3.3-70B-Instruct & 70B\\
        \midrule
        Mistral-7B     & mistralai/ \\&Mistral-7B-Instruct-v0.3 & 7B\\
        \midrule
        Phi3-medium  & microsoft/ \\&Phi-3-medium-4k-instruct & 14B\\
        \midrule
        GPT-4o mini     & gpt-4o-mini-2024-07-18 & --\\
        \midrule
        o3-mini     & o3-mini-2025-01-31 & --\\ 
        \midrule
        BERT-base & google-bert/bert-base-uncased & 110M\\
        \midrule
        ModernBERT & answerdotai/ModernBERT-base & 150M\\
        \bottomrule
    \end{tabular}
    \caption{Model checkpoints from HuggingFace library and OpenAI API.}
    \label{tab:model_checkpoints}
\end{table}

We use AI assistants to assist with minor debugging and coding.

\subsection{Prompts for Experiments}
\label{sec:prompts-for-experiments}

In this section, we provide the templates and prompts used for zero-shot, one-shot, and three-shot settings.

\subsubsection{Templates}
For zero-shot inference, we use the following:

\begin{mybox}[Zero-Shot Template]
\#\#\#SYSTEM: \textit{[System Prompt]}

\#\#\#PROMPT: \textit{[Unreliability Definition]}

\textit{[Shot Instruction]}

\#\#\#INPUT: \textit{[Narrative]}

\#\#\#SOLUTION: 
\end{mybox}

For few-shot inference, we include shots after the shot instruction. We notice best performance when we repeat the shot instruction a second time after the shots. We use the following:

\begin{mybox}[Few-Shot Template]
    
\#\#\#SYSTEM: \textit{[System Prompt]}

\#\#\#PROMPT: \textit{[Unreliability Definition]}

\textit{[Shot Instruction]}

\textit{[Shots]}

\textit{[Shot Instruction]}

\#\#\#INPUT: \textit{[Narrative]}

\#\#\#SOLUTION: 
\end{mybox}

\subsubsection{System Prompts}
We replace \textit{[System Prompt]} in the templates with the corresponding system prompt: \\

\noindent\underline{\textbf{Intra-narrational}}: 

\noindent Determine if the INPUT narrative contains any verbal tics.

\noindent\underline{\textbf{Inter-narrational}}: 

\noindent Determine if the narrator in the INPUT narrative is unreliable based on another character's point of view.

\noindent\underline{\textbf{Inter-textual}}: 

\noindent Determine if the narrator in the INPUT narrative fits a character trope.

\subsubsection{Definitions}
We replace \textit{[Unreliability Definition]} in the Template with the corresponding set of definitions: \\

\noindent\underline{\textbf{Intra-narrational}}:

\noindent Here is a list of verbal tics:  

\begin{itemize}[topsep=1pt, leftmargin=*, noitemsep]
    \item Admission of fault/bias: Narrator directly states that he/she has made mistakes, has particular biases, does not know all the details, or is reporting information from another character who is likely unreliable. Examples: ``Like others of my generation...'', ``I tend to see things from a particular point of view'' 
    \item Defensive tone: Narrator uses multiple phrases in protestation. Examples: ``Let me make perfectly clear'', ``I should say'', ``I should point out'', ``let me make it immediately clear'', ``I feel I should explain''
    \item Digressions: Narrator veers off-topic at least once. Examples: ``I will do that in a minute. By the way,...''
    \item Hedging language: Narrator uses phrases that indicate uncertainty or vagueness. Examples: ``it seems that'', ``it appears to be'', ``I think'', ``maybe'', ``sort of''
    \item Inconsistencies: Narrator gives contradicting statements and/or events don't add up. ``I am a nobody. But look! There is a plane drawing my name in the sky.''
    \item Selective Memory: Narrator acknowledges they may have forgotten important details or their memory is faulty. Examples: ``It was so long ago, it's hard to remember'', ``My memory is not what it used to be''
    \item Statement of Potential Disbelief: Narrator acknowledges the narrative may sound unlikely. Examples: ``You might not believe me, but'', ``what happened next might seem strange''
\end{itemize}

Given this list of verbal tics and a narrative, we define the following options:

\begin{enumerate}[topsep=1pt, leftmargin=1.5cm, noitemsep]
    \item[<A>:] Verbal tics present in narrative
    \item[<B>:] No verbal tics present in narrative
\end{enumerate}

\noindent\underline{\textbf{Inter-narrational}}: 

\noindent We define the following types of unreliable narrators based on an alternative character's point of view:

\begin{enumerate}[topsep=1pt, leftmargin=1.5cm, noitemsep]
    \item [<A>:] Same unreliable narrator over time: Narrator is reflecting on his/her past self, who is unreliable, AND the present-day narrator does not indicate change within narrative snippet.
    \item [<B>:] Other character contradiction: Another character contradicts narrator who has demonstrated at least one form of intra-narrational unreliability. 
    \item [<C>:] Neither: narrator is reliable from different characters' points of view.
\end{enumerate}

\noindent\underline{\textbf{Inter-textual}}: 

\noindent We define the following types of unreliable narrators based on character tropes:

\begin{enumerate}[topsep=1pt, leftmargin=1.5cm, noitemsep]
    \item[<A>:] Naïf: Narrator who is a foil to a lamentable social condition and who lacks experience to fully understand the narrated events or the complexity of their environment. (Defining characteristic: Blind to wrongs) 
    \item [<B>:] Madman: Narrator, often with a frantic voice, who feels deep positive or negative emotions toward others and is maddened by perceived torture or alienation. (Defining characteristic: Highly emotional)
    \item [<C>:] Pícaro: Socially aware rogue or antihero who experiences the rise and fall of fortune while attempting to improve their prospects and cunningly justify their chaotic worldview. (Defining characteristic: Tries to be clever)
    \item [<D>:] Clown: Narrator who offers reinterpretations that repackage internal and/or external conflict in a new light, potentially from behind a facade that allows them to say whatever they want. (Defining characteristic: Flips the narrative)
    \item [<E>:] None: Narrator is reliable based on character tropes.
\end{enumerate}

\subsubsection{Instruction}
We replace \texttt{[Shot Instruction]} in the Template with the corresponding instruction: 

\noindent \underline{\textbf{Intra-narrational:}} 
Does the narrative demonstrate any verbal tics? DO NOT GIVE ANY DESCRIPTION. Generate ONLY THE CORRESPONDING LETTER enclosed in angle brackets ('<A>' or '<B>').

\noindent \underline{\textbf{Inter-narrational}:}
What type of narrator is given? DO NOT GIVE ANY DESCRIPTION. Generate ONLY THE CORRESPONDING LETTER enclosed in angle brackets ([List of possible letters enclosed in angle brackets]).

\noindent \underline{\textbf{Inter-textual}:}
What type of narrator is given? DO NOT GIVE ANY DESCRIPTION. Generate ONLY THE CORRESPONDING LETTER enclosed in angle brackets ('<A>' or '<B>' or '<C>' or '<D>' or '<E>').

\subsubsection{Shots}

In this section we show one example shot per label. During inference, we replace \textit{[Shots]} in the Template with one or three shots per label: \\

\noindent \underline{\textbf{Intra-narrational:}} 

INPUT 1: This is written from memory, unfortunately. If I could have brought with me the material I so carefully prepared, this would be a very different story. Whole books full of notes, carefully copied records, firsthand descriptions, and the pictures—that’s the worst loss. We had some bird’s-eyes of the cities and parks; a lot of lovely views of streets, of buildings, outside and in, and some of those gorgeous gardens, and, most important of all, of the women themselves.
Nobody will ever believe how they looked. Descriptions aren’t any good when it comes to women, and I never was good at descriptions anyhow. But it’s got to be done somehow; the rest of the world needs to know about that country.

SOLUTION 1: <A>

INPUT 2: It was a wild night. Driving clouds kept hiding and revealing the stormy-looking moon. I was out-of-doors. I could not remain in the house; it had felt too small for me, but now nature felt too large. I dimly saw the huge pile of the schloss defined against the gray light; sometimes when the moon unveiled herself it started out clear, and black, and grim. I heard the wind moan among the trees, heard the great dogs baying from the kennels; from an open window came rich, low, mellow sounds. Old Brunken was in the music-room, playing to himself upon the violoncello.

SOLUTION 2: <B>\\

\noindent \underline{\textbf{Inter-narrational}:}
INPUT 1: In due time I found my ghost, or ghosts rather, for there were two of them. Up till that hour I had sympathized with Mr. Besant’s method of handling them, as shown in “The Strange Case of Mr. Lucraft and Other Stories.” I am now in the Opposition.
We will call the bungalow Katmal dâk-bungalow. But THAT was the smallest part of the horror. A man with a sensitive hide has no right to sleep in dâk-bungalows. He should marry. Katmal dâk-bungalow was old and rotten and unrepaired. The floor was of worn brick, the walls were filthy, and the windows were nearly black with grime. It stood on a bypath largely used by native Sub-Deputy Assistants of all kinds, from Finance to Forests; but real Sahibs were rare. The khansamah, who was nearly bent double with old age, said so.

SOLUTION 1: <A> 

INPUT 2: “I suppose you would like to take them to the Casino to play roulette? Well, excuse my speaking so plainly, but I know how addicted you are to gambling. Though I am not your mentor, nor wish to be, at least I have a right to require that you shall not actually compromise me.”
“I have no money for gambling,” I quietly replied.
“But you will soon be in receipt of some,” retorted the General, reddening a little as he dived into his writing desk and applied himself to a memorandum book. From it he saw that he had 120 roubles of mine in his keeping.
“Let us calculate,” he went on. “We must translate these roubles into thalers. Here—take 100 thalers, as a round sum. The rest will be safe in my hands.”
In silence I took the money.

SOLUTION 2: <B>

INPUT 3: “I heard the sound of a stick and a shambling step on the flags in the passage outside, and the door creaked on its hinges as a second old man entered, more bent, more wrinkled, more aged even than the first. He supported himself by a single crutch, his eyes were covered by a shade, and his lower lip, half-averted, hung pale and pink from his decaying yellow teeth. He made straight for an arm-chair on the opposite side of the table, sat down clumsily, and began to cough. The man with the withered arm gave this new-comer a short glance of positive dislike; the old woman took no notice of his arrival, but remained with her eyes fixed steadily on the fire.”

SOLUTION 3: <C>\\

\noindent \underline{\textbf{Inter-textual}:}
INPUT 1: They went off and I got aboard the raft, feeling bad and low, because I knowed very well I had done wrong, and I see it warn’t no use for me to try to learn to do right; a body that don’t get started right when he’s little ain’t got no show—when the pinch comes there ain’t nothing to back him up and keep him to his work, and so he gets beat. Then I thought a minute, and says to myself, hold on; s’pose you’d a done right and give Jim up, would you felt better than what you do now? No, says I, I’d feel bad—I’d feel just the same way I do now. Well, then, says I, what’s the use you learning to do right when it’s troublesome to do right and ain’t no trouble to do wrong, and the wages is just the same? I was stuck. I couldn’t answer that. So I reckoned I wouldn’t bother no more about it, but after this always do whichever come handiest at the time.

SOLUTION 1: <A>

INPUT 2: You lie, cursed dog! What a scandalous tongue! As if I did not know that it is envy which prompts you, and that here there is treachery at work—yes, the treachery of the chief clerk. This man hates me implacably; he has plotted against me, he is always seeking to injure me. I'll look through one more letter; perhaps it will make the matter clearer.

SOLUTION 2: <B>

INPUT 3: I grew the greatest artist of my time and worked myself out of every danger with such dexterity, that when several more of my comrades ran themselves into Newgate presently, and by that time they had been half a year at the trade, I had now practised upwards of five years, and the people at Newgate did not so much as know me; they had heard much of me indeed, and often expected me there, but I always got off, though many times in the extremest danger.

SOLUTION 3: <C>

INPUT 4: Therefore, my dear friend and companion, if you should think me somewhat sparing of my narrative on my first setting out—bear with me,—and let me go on, and tell my story my own way:—Or, if I should seem now and then to trifle upon the road,—or should sometimes put on a fool’s cap with a bell to it, for a moment or two as we pass along,—don’t fly off,—but rather courteously give me credit for a little more wisdom than appears upon my outside;—and as we jog on, either laugh with me, or at me, or in short do any thing,—only keep your temper.

SOLUTION 4: <D>

INPUT 5: I first heard of Ántonia on what seemed to me an interminable journey across the great midland plain of North America. I was ten years old then; I had lost both my father and mother within a year, and my Virginia relatives were sending me out to my grandparents, who lived in Nebraska. I travelled in the care of a mountain boy, Jake Marpole, one of the ‘hands’ on my father’s old farm under the Blue Ridge, who was now going West to work for my grandfather. Jake’s experience of the world was not much wider.

SOLUTION 5: <E>

\section{Additional Details About Classification Results}
\label{sec:all-model-results}

We show class-wise performance for combined test domains in Table \ref{tab:class-wise} and performance breakdowns for each textual domain (akin to Table \ref{tab:f1scores-breakdown}) using the remaining LLMs: Llama3.3-70B (Table~\ref{tab:f1scores-breakdown-llama3.3-70b}), Mistral-7B (Table~\ref{tab:f1scores-breakdown-mistral}), Phi3-medium (Table~\ref{tab:f1scores-breakdown-phi}), GPT-4o mini (Table~\ref{tab:f1scores-breakdown-gpt4o-mini}), and o3-mini (Table~\ref{tab:f1scores-breakdown-gpt-o3-mini}). Furthermore, we provide an error analysis of the classification tasks (Appendix~\ref{sec:errors-classifying-un}), examples showing how different tasks have different complexities (Appendix~\ref{sec:demonstration-of-task-complexities}), and examples of how LLM learn from snippets to make better predictions (Appendix~\ref{sec:effect-of-learning-from-examples}).

\subsection{Errors Classifying Unreliable Narrators}
\label{sec:errors-classifying-un}
We analyze the LLM classifications by generating explanations (i.e., we prompt the LLMs to explain their choices) and make the following observations about incorrectly classified unreliable narrators.

For intra-narrational unreliability, the LLM over-predicts the presence of hedging language and defensive language and is overly-sensitive to phrases that annotators do not classify as either verbal tic. The LLM also struggles to recognize digressions. In some cases, the LLM states a verbal tic but mistakenly determines that it is not strong enough to classify as an unreliability.

For inter-narrational unreliability, the LLM tends to overpredict “other character contradiction” by anticipating another character might contradict the narrator without explicit evidence within the context. Sometimes the LLM considers instances where the narrator contradicts themself or seems manipulative as “other character contradiction.”

For inter-textual unreliability, the LLM struggles with ambiguous samples (i.e., samples that contain characteristics of multiple types of unreliabilities). For these samples, the LLM seems to over-emphasize characteristics of less dominant unreliabilities. For example, the LLM often over-predicts Madman for samples where the narrator describes emotions. On the other hand, the LLM sometimes ignores characteristics of a Madman noted by the annotators if the LLM describes other minor attributes corresponding to another unreliability.

\begin{table*}[t]
\centering
\footnotesize
\begin{tabular}{llccccc}
\toprule
\texttt{Llama3.1-8B}&& \textbf{(A)}& \textbf{(B)}& \textbf{(C)}& \textbf{(D)}& \textbf{(E)}\\
\toprule
\textbf{CL}    & \textit{Intra-nar}  & 69.67$\pm$1.02 & 50.30$\pm$1.54 &--&--&--\\
      & \textit{Inter-nar}  & 2.90$\pm$1.25  & 39.36$\pm$2.16 & 78.21$\pm$0.82 &--&--\\
      & \textit{Inter-tex}  & 10.96$\pm$2.17 & 14.90$\pm$1.98 & 15.66$\pm$2.50 & 11.84$\pm$1.91 & 57.24$\pm$1.18 \\
\midrule
\textbf{FT}    & \textit{Intra-nar}  & 59.84$\pm$1.18 & 48.87$\pm$1.33 &--&--&--\\
      & \textit{Inter-nar}  & 6.89$\pm$1.86  & 20.00$\pm$2.18 & 80.16$\pm$0.75 &--&--\\
      & \textit{Inter-tex}  & 4.49$\pm$1.70  & 16.08$\pm$1.93 & 10.07$\pm$2.28 & 16.81$\pm$2.13 & 59.59$\pm$1.12 \\
\midrule
\textbf{Zero-shot} & \textit{Intra-nar} & 29.77$\pm$1.40 & 5.80$\pm$1.22  &--&--&--\\
      & \textit{Inter-nar}  & 6.73$\pm$1.74  & 20.69$\pm$2.05 & 10.46$\pm$1.03 &--&--\\
      & \textit{Inter-tex}  & 12.75$\pm$1.98 & 2.47$\pm$1.01  & 4.68$\pm$1.87  & 1.62$\pm$0.92  & 16.43$\pm$1.55 \\
\midrule
\textbf{One-shot} & \textit{Intra-nar} & 29.56$\pm$1.37 & 7.38$\pm$1.35  &--&--&--\\
      & \textit{Inter-nar}  & 9.26$\pm$1.83  & 16.23$\pm$2.03 & 11.99$\pm$1.13 &--&--\\
      & \textit{Inter-tex}  & 16.57$\pm$2.38 & 10.25$\pm$1.87 & 3.37$\pm$1.68  & 4.52$\pm$1.44  & 20.31$\pm$1.72 \\
\midrule
\textbf{Three-shot} & \textit{Intra-nar} & 30.03$\pm$1.34 & 4.52$\pm$1.09  &--&--&--\\
      & inter-nar  & 10.29$\pm$1.98 & 15.66$\pm$2.02 & 12.84$\pm$1.14 &--&--\\
      & inter-tex  & 17.41$\pm$2.21 & 2.52$\pm$1.01  & 9.02$\pm$2.38  & 2.17$\pm$1.06  & 14.75$\pm$1.52 \\ 
\bottomrule
\end{tabular}
\caption{Class-wise F1 (macro) scores for combined test domains for Llama3.1-8B. For \textit{intra-nar}: \textbf{(A)} $\rightarrow$ verbal tics, \textbf{(B)} $\rightarrow$ none. For \textit{inter-nar}: \textbf{(A)} $\rightarrow$ ``Same unreliable character over time'', \textbf{(B)} $\rightarrow$ ``Other character contradiction'', \textbf{(C)} $\rightarrow$ none. For \textit{inter-tex} unreliabilities: \textbf{(A)} $\rightarrow$ Naïf, \textbf{(B)} $\rightarrow$ Madman, \textbf{(C)} $\rightarrow$ Pícaro, \textbf{(D)} $\rightarrow$ Clown, \textbf{(C)} $\rightarrow$ none. 
We make the following observations:  For \textit{intra-nar}, CL and FT methods improve the performance of both classes. For \textit{inter-nar}, multi-shot setting improves performance for ‘same unreliable narrator over time’ and Reliable; CL and FT improves ‘other character contradiction’ and Reliable. For \textit{inter-tex}, CL and FT substantially improves performance for Madman, Picaro, Clown, and Reliable.}
\label{tab:class-wise}
\end{table*}

\begin{table*}[t]
    \centering
    \footnotesize
    \begin{tabular}{llcc||ccc}
        \toprule
        \texttt{Llama3.3-70B}&& \textbf{CL} & \textbf{Fine-tuned} & \textbf{Zero-Shot} & \textbf{One-Shot} & \textbf{Three-Shot}\\
        \toprule
        \textbf{Intra-nar}  &\hspace{.25cm}\textit{Fiction}      &54.74±2.03	&54.72±1.96	&45.90±1.72	&64.56±1.92	&62.94±2.14 \\
        &\hspace{.25cm}\textit{Blog post}    &49.34±2.22	&49.48±2.13	&53.04±2.33	&64.65±2.26	&62.74±2.22 \\ 
        &\hspace{.25cm}\textit{Subreddit}    &44.55±1.96	&44.54±2.00	&46.37±0.42	&56.44±2.94	&47.16±1.26 \\
        &\hspace{.25cm}\textit{Review}       &56.39±2.26	&56.37±2.26	&71.51±2.11	&69.91±2.01	&72.82±2.01 \\        
        \cmidrule{2-7}
        \textbf{Inter-nar}  &\hspace{.25cm}\textit{Fiction} &43.47±3.21	&35.81±2.39	&27.48±2.41	&23.40±1.58	&24.23±1.72 \\
        &\hspace{.25cm}\textit{Blog post}    &30.61±1.95	&25.68±0.68	&26.70±1.22	&40.72±2.01	&45.74±2.78 \\ 
        &\hspace{.25cm}\textit{Subreddit}    &28.57±3.50	&27.84±1.88	&28.27±1.32	&31.38±1.90	&28.29±1.81 \\
        &\hspace{.25cm}\textit{Review}       &31.32±0.57	&31.94±0.22	&33.98±1.59	&29.44±1.41	&37.84±2.60 \\
        
        \cmidrule{2-7}
        \textbf{Inter-tex} &\hspace{.25cm}\textit{Fiction}      &23.45±1.75	&23.42±1.67	&29.83±1.91	&32.59±1.90	&31.95±1.83 \\
        &\hspace{.25cm}\textit{Blog post}    &24.63±1.83	&24.61±1.76	&35.64±2.29	&42.54±2.39	&33.63±2.05 \\ 
        &\hspace{.25cm}\textit{Subreddit}    &13.43±1.38	&13.46±1.40	&19.68±2.32	&27.53±1.99	&19.40±1.47 \\
        &\hspace{.25cm}\textit{Review}       &22.66±1.81	&22.61±1.75	&28.94±1.32	&20.55±0.97	&27.93±2.20 \\

        \bottomrule
    \end{tabular}
    \caption{Breakdown of unreliability F1 (macro) scores for each domain using Llama3.3-70B.}
    \label{tab:f1scores-breakdown-llama3.3-70b}
\end{table*}

\begin{table*}[t]
    \centering
    \footnotesize
    \begin{tabular}{llcc||ccc}
        \toprule
        \texttt{Mistral}&& \textbf{CL} & \textbf{Fine-tuned} & \textbf{Zero-Shot} & \textbf{One-Shot} & \textbf{Three-Shot}\\
        \toprule
        \textbf{Intra-nar}  &\hspace{.25cm}\textit{Fiction}      &54.66±2.06	&56.76±2.04	&54.91±2.13	&49.74±1.98	&53.36±2.08 \\
        &\hspace{.25cm}\textit{Blog post}    &54.21±2.19	&56.79±2.17	&54.89±2.19	&54.19±2.29	&49.34±2.20 \\ 
        &\hspace{.25cm}\textit{Subreddit}    &46.43±0.41	&49.92±2.12	&48.83±1.80	&47.20±1.33	&51.54±2.51 \\
        &\hspace{.25cm}\textit{Review}       &67.73±2.14    &62.35±2.13 &68.54±2.06	&52.34±2.24	&57.73±2.17 \\        
        
        \midrule
        \textbf{Inter-nar}  &\hspace{.25cm}\textit{Fiction} &22.16±1.02	&31.37±0.24	&12.28±1.38	&30.79±2.01	&28.97±1.87 \\
        &\hspace{.25cm}\textit{Blog post}    &38.73±1.42	&25.07±0.54	&18.63±1.60	&35.70±1.93	&33.06±1.63 \\ 
        &\hspace{.25cm}\textit{Subreddit}    &33.14±1.92	&14.41±0.78	&25.66±1.15	&37.88±2.18	&30.96±1.99 \\
        &\hspace{.25cm}\textit{Review}       &30.58±1.44	&32.12±0.20	&21.38±1.30	&27.91±1.55	&32.18±1.95 \\
        
        \midrule
        \textbf{Inter-tex} &\hspace{.25cm}\textit{Fiction}      &32.43±2.02	&29.89±1.69	&21.76±1.75	&19.35±1.56	&18.80±1.49 \\
        &\hspace{.25cm}\textit{Blog post}    &36.59±2.25	&28.15±1.75	&22.74±1.70	&22.28±1.82	&19.29±1.65 \\ 
        &\hspace{.25cm}\textit{Subreddit}    &22.53±1.80	&9.08±0.92	&15.78±1.50	&16.03±1.49	&13.29±1.27 \\
        &\hspace{.25cm}\textit{Review}       &27.15±1.98	&30.40±0.82	&20.63±1.08	&15.73±0.84	&17.11±0.97 \\
        \bottomrule
        
    \end{tabular}
    \caption{Breakdown of unreliability F1 (macro) scores for each domain for Mistral-7B.}
    \label{tab:f1scores-breakdown-mistral}
\end{table*}

\begin{table*}[t]
    \centering
    \footnotesize
    \begin{tabular}{llcc||ccc}
        \toprule
        \texttt{Phi}&& \textbf{CL} & \textbf{Fine-tuned} & \textbf{Zero-Shot} & \textbf{One-Shot} & \textbf{Three-Shot}\\
        \toprule
        \textbf{Intra-nar}  &\hspace{.25cm}\textit{Fiction}      &64.14±1.93	&43.53±1.88	&62.56±2.04	&46.56±1.86	&45.91±1.76	\\
        &\hspace{.25cm}\textit{Blog post}    &53.42±2.23	&36.93±2.46	&55.37±2.25	&43.71±1.52	&46.31±1.88	\\ 
        &\hspace{.25cm}\textit{Subreddit}    &51.18±2.37	&59.34±3.03	&51.82±2.54	&47.71±1.37	&46.26±0.42 \\
        &\hspace{.25cm}\textit{Review}       &46.26±2.03	&68.94±2.07	&70.25±2.06	&40.82±2.01	&40.97±1.89	\\        
        
        \midrule
        \textbf{Inter-nar}  &\hspace{.25cm}\textit{Fiction} &29.41±2.30	&46.54±2.75	&18.45±1.55	&13.65±1.45	&16.08±1.39 \\
        &\hspace{.25cm}\textit{Blog post}    &13.34±1.17	&28.77±1.24	&26.81±1.74	&26.55±1.83	&26.41±1.76	\\ 
        &\hspace{.25cm}\textit{Subreddit}    &23.62±1.18	&31.35±1.41	&25.16±1.12	&28.53±1.81	&27.93±1.69 \\
        &\hspace{.25cm}\textit{Review}       &22.89±1.32	&36.38±1.84	&30.49±3.10	&24.94±1.73	&28.20±2.06	\\
        
        \midrule
        \textbf{Inter-tex} &\hspace{.25cm}\textit{Fiction}      &27.24±1.63	&34.01±1.96	&23.90±1.44	&14.39±1.39	&13.55±1.43	\\
        &\hspace{.25cm}\textit{Blog post}    &26.62±1.40	&30.90±2.11	&28.42±1.78	&20.37±1.68	&18.38±1.42 \\ 
        &\hspace{.25cm}\textit{Subreddit}    &17.94±1.31	&16.36±1.40	&22.21±1.65	&18.36±1.60	&15.90±1.54	\\
        &\hspace{.25cm}\textit{Review}       &28.20±1.70	&23.69±1.80	&35.70±1.92	&22.23±2.69	&17.78±1.11 \\

        \bottomrule
        
    \end{tabular}
    \caption{Breakdown of unreliability F1 (macro) scores for each domain using Phi3-medium.}
    \label{tab:f1scores-breakdown-phi}
\end{table*}

\begin{table*}[t]
    \centering
    \footnotesize
    \begin{tabular}{llccc}
        \toprule
        \texttt{GPT-4omini}&& \textbf{Zero-Shot} & \textbf{One-Shot} & \textbf{Three-Shot}\\
        \toprule
        \textbf{Intra-nar}  &\hspace{.25cm}\textit{Fiction}	&45.48±1.90	&56.29±2.03	&54.66±2.11\\
        &\hspace{.25cm}\textit{Blog post}    &55.03±2.20	&46.65±1.90	&44.40±1.74\\ 
        &\hspace{.25cm}\textit{Subreddit}    &49.33±2.14	&46.97±0.38	&54.71±2.85\\
        &\hspace{.25cm}\textit{Review}       &41.67±1.96	&52.12±2.37	&53.31±2.29\\        
        
        \midrule
        \textbf{Inter-nar}  &\hspace{.25cm}\textit{Fiction} &33.27±2.62	&38.38±2.77	&23.95±2.18\\
        &\hspace{.25cm}\textit{Blog post}    &23.81±1.23	&25.42±1.28	&22.14±1.05\\ 
        &\hspace{.25cm}\textit{Subreddit}    &27.35±1.18	&30.75±1.43	&30.51±1.50\\
        &\hspace{.25cm}\textit{Review}       &28.15±0.94	&31.46±1.31	&27.41±1.34\\
        
        \midrule
        \textbf{Inter-tex} &\hspace{.25cm}\textit{Fiction}  &21.07±1.51	&28.00±1.95	&26.22±1.86\\
        &\hspace{.25cm}\textit{Blog post}    &21.66±1.77	&29.06±2.09	&24.21±1.99\\ 
        &\hspace{.25cm}\textit{Subreddit}    &11.90±1.17	&9.81±1.31	&13.81±1.35\\
        &\hspace{.25cm}\textit{Review}       &16.73±1.17	&15.75±0.32	&15.69±0.32\\
        \bottomrule
        
    \end{tabular}
    \caption{Breakdown of unreliability F1 (macro) scores for each domain using GPT-4o mini.}
    \label{tab:f1scores-breakdown-gpt4o-mini}
\end{table*}

\begin{table*}[t]
    \centering
    \footnotesize
    \begin{tabular}{llccc}
        \toprule
        &&\textbf{Zero-Shot} & \textbf{One-Shot} & \textbf{Three-Shot}\\
        \toprule
        \textbf{Intra-nar}  &\hspace{.25cm}\textit{Fiction}      &41.84±1.93	&45.07±1.96	&43.08±1.97 \\
        &\hspace{.25cm}\textit{Blog post}    &47.28±2.23	&47.45±2.15	&49.60±2.18\\ 
        &\hspace{.25cm}\textit{Subreddit}    &39.87±1.79	&42.12±1.94	&42.70±1.97\\
        &\hspace{.25cm}\textit{Review}       &39.89±1.91	&39.25±1.96	&41.88±2.03\\        
        
        \midrule
        \textbf{Inter-nar}  &\hspace{.25cm}\textit{Fiction} &35.13±1.84	&32.33±1.42	&31.57±1.19 \\
        &\hspace{.25cm}\textit{Blog post}    &27.20±1.63	&23.83±0.62	&24.94±1.25	\\ 
        &\hspace{.25cm}\textit{Subreddit}    &30.35±1.43	&26.87±1.40	&26.39±1.44	\\
        &\hspace{.25cm}\textit{Review}       &36.02±2.71	&32.12±0.20	&32.16±0.20	\\
        
        \midrule
        \textbf{Inter-tex} &\hspace{.25cm}\textit{Fiction}      &21.99±1.73	&22.81±1.68	&22.85±1.85	\\
        &\hspace{.25cm}\textit{Blog post}    &21.66±1.72	&16.16±1.37	&22.48±1.71	\\ 
        &\hspace{.25cm}\textit{Subreddit}    &13.88±1.69	&14.13±1.55	&16.13±1.65	\\
        &\hspace{.25cm}\textit{Review}       &16.65±1.14	&15.44±0.33	&17.84±1.54	\\
        \bottomrule
        
    \end{tabular}
    \caption{Breakdown of unreliability F1 (macro) scores for each domain using o3-mini.}
    \label{tab:f1scores-breakdown-gpt-o3-mini}
\end{table*}

\subsection{Demonstration of Task Complexities}
\label{sec:demonstration-of-task-complexities}

The following example demonstrates how the intra-narrational task is more easily predicted by analyzing relatively explicit instances of verbal tics than the inter-textual task. We note that the inter-textual task is more challenging because it requires a deeper understanding of the narrator's frame-of-thought and situation. Model results are from the CL method using Llama3.1-8B.\\

\textit{Blog post:} ``I got pulled over this morning!! Err... Yesterday morning... On my way back to Owings Mills for my sorority retreat deal. Yeah. Totally didn't notice the cops like I usually do, and apparently flew past him doing 85. And I didn't slow down when I got to 695 (they really need to up the speed limit to 65 over there I swear), so he asked me what my excuse was. 'I don't have one Sir.' 'License and registration please.' I sit there, waiting for the ticket. He hands me everything back and says Slow down. That's way too fast. Have a good day.' '... You too officer.' Yeah. So never going over 65 on 695 ever again. Cruise control... Buddy... how you doin? Told mom and dad I was doing 70 in a 55 they said if I get a ticket, I'll more than likely not have car insurance. Gah. Either way, no more than 10 above. Doing the SPEED limit on the highways around here is dangerous man. People almost hit me like, three times on 695.''\\

\textit{Intra-narrational Prediction:} A

\textit{Intra-narrational Gold Label:} A

\textit{Why?} The model easily finds at least one verbal tic (e.g., inconsistency in the first two sentences, hedging language, admission of fault/bias) in the narrative.\\

\textit{Inter-textual Prediction:} E

\textit{Inter-textual Gold Label:} D

\textit{Why?} The model misses contextual information scattered throughout the narrative that indicates the narrator is bouncing between conflicting interpretations of getting a speeding ticket.

\subsection{Effect of Learning from Examples}
\label{sec:effect-of-learning-from-examples}

In this section, we show one sample for each type of unreliability where the model predicts an incorrect label in the zero-shot setting, and the model learns from examples in the few-shot setting to predict the correct label.\\

\noindent\underline{\textbf{Intra-narrational:}} \textit{{(Subreddit)}}
``I’m (21M) a junior at an Ivy League school that gets really into holidays, and the student social committee spends a ton of money on throwing Halloween events.
  
In particular, there’s going to be a massive, very expensive-to-host squid game party. It’s going to be super fancy with squid game consumes and real life-size games and awesome food, and the budget is huge. 
 
I asked for a much smaller amount for a simple masquerade party my club is throwing. Everyone is extremely annoyed with me for “siphoning committee funds” away from the squid game party. But I barely asked for anything compared to what they’re spending. I don’t see the big deal. My friend planning the squid game party is particularly irate.''

\textit{0-Shot Prediction:} B

\textit{1-Shot Prediction:} A

\textit{Gold Label:} A

\textit{Why?} There are many examples of hedging language: “gets really into holidays”, “spends a ton of money”, “going to be a massive, very expensive”, “super fancy”, “much smaller amount”\\

\noindent\underline{\textbf{Inter-narrational:}} \textit{{(Review)}}
``I stayed at the Monaco-Chicago back in April. I was in town on business, and the hotel was recommended by a friend of mine. Having spent a weekend there, I have no idea what my friend was talking about. The complimentary morning coffee was weak; the fitness room was dimly lit; and I thought I'd have to have my clothes mailed back to me when I used their supposed 'overnight' laundry service for a suit I spilled some wine on. My room was adequate, but nowhere near what I've seen elsewhere at this price point. Recent renovation must be slang for 'everything is stiff and smells of industrial adhesive.' The mattress in my room was incredibly firm, and I slept poorly. When I travel, I expect an experience similar to or better than my experience at home. At most hotels, I receive excellent service and comfortable accomodations. This was an exception to my usual, and I won't be back anytime soon.''

\textit{0-Shot Prediction:} B

\textit{1-Shot Prediction:} A

\textit{Gold Label:} B

\textit{Why?} The friend who recommended the hotel is in contradiction with the narrator.\\

\noindent\underline{\textbf{Inter-textual:}} \textit{{(Blog post)}} 
``So last week, my 95 year old Grandfather fell and cracked his vertebrae.. Long story short, it was a rough road last week with dr.s saying that he was lucky to be alive and today I receive a phone call from my sister saying that his kidneys had failed and that my Grandmother and aunts, uncles and father had made the decision to pull his life support (he has been unresponsive a day after it happened). While we live 6 hours away now - I used to live in the same town as all of my family until I went away to college.. EVERY Sunday was spent at my Grandparents house and I have so many memories of him when I was a kid. But I think the hardest part for me is my thinking of my Grandmother and I wonder, How do you do it? How do you say goodbye? How do you kiss your loves lips for the last time and know that you will never be able to do that again? How do you share your whole life with someone go through wars, 6 kids and over 50 years and then in one moment it is all gone. My prayers are with her because I don't know how she can do this''

\noindent\textit{0-Shot Prediction:} E

\noindent\textit{1-Shot Prediction:} B

\noindent\textit{Gold Label:} B

\noindent\textit{Why?} Narrator is a madman because of the strong emotions described at the loss of the grandfather. Narrator seems to feel deep positive emotions for the grandparents

\section{Details of Analysis}
\label{sec:details_of_analysis}

In this section, we give additional information about how we automatically determine properties of the narratives for the analysis in Section \ref{sec:analysis}. We give all prompts used for inferring the properties of narratives (Section~\ref{sec:analysis-prompts}) and additional analysis results for Llama3.3-70B, Mistral-7B, and Phi3-medium (Section~\ref{sec:additional-analysis-results}).

\subsection{Prompts}
\label{sec:analysis-prompts}
For our analysis experiments, we prompt Llama3.3-70B
with the following:

\begin{mybox}[Analysis Template]
\#\#\#PROMPT: Consider the narrative. 

\textit{[RQ\#]}

\#\#\#INPUT: \textit{[Narrative]}

\#\#\#SOLUTION: Final label:
\end{mybox}

For each research question, we replace \textit{[RQ\#]} in the Analysis Template with one of the following:

\begin{itemize}
    \item \textit{RQ1}: Is the narrator female, male, other, or ambiguous? STATE ONLY THE GENDER AS A LABEL.

    \item \textit{RQ2}: Is the style of narration “conversational” or “descriptive”? For “conversational” the narrator is chatty to the reader. For “descriptive” the narrator primarily describes the setting or situation. STATE ONLY THE STYLE AS A LABEL.

    \item \textit{RQ3}: Is the narration tone primarily “positive” or “negative” or “neutral”? STATE ONLY THE TONE AS A LABEL.

    \item \textit{RQ4}: Including the narrator, how many explicit characters play a role in the narrative? STATE ONLY THE NUMBER OF CHARACTERS AS AN INTEGER LABEL.
\end{itemize}

\subsection{Errors Prompting for Analysis}

\label{sec:prompting-for-analysis}
We hand-verify 200 narratives to ensure the inferred properties of narratives are comparable to human judgement. We observe fewer than 15 misclassifications (minimum accuracy of 92.50\%) for each task and notice the following.

For gender, some narrators with ambiguous genders are incorrectly predicted Male or Female due to context clues creating bias for that gender (e.g., ambiguous narrator ironing a suit is predicted Male, or ambiguous narrator getting mani/pedi is predicted Female).

For style, in general, if a narrative begins with description of the narrator or if the sample contains obvious verbal tics (e.g., multiple examples of hedging language), the sample is typically classified by the LLM as conversational.

For sentiment, sometimes the LLM predicts the wrong sentiment if the narrator speaks sarcastically.

\subsection{Analysis Results with Other Models}
\label{sec:additional-analysis-results}
\begin{figure}[t]
    \centering
    \includegraphics[width=\linewidth]{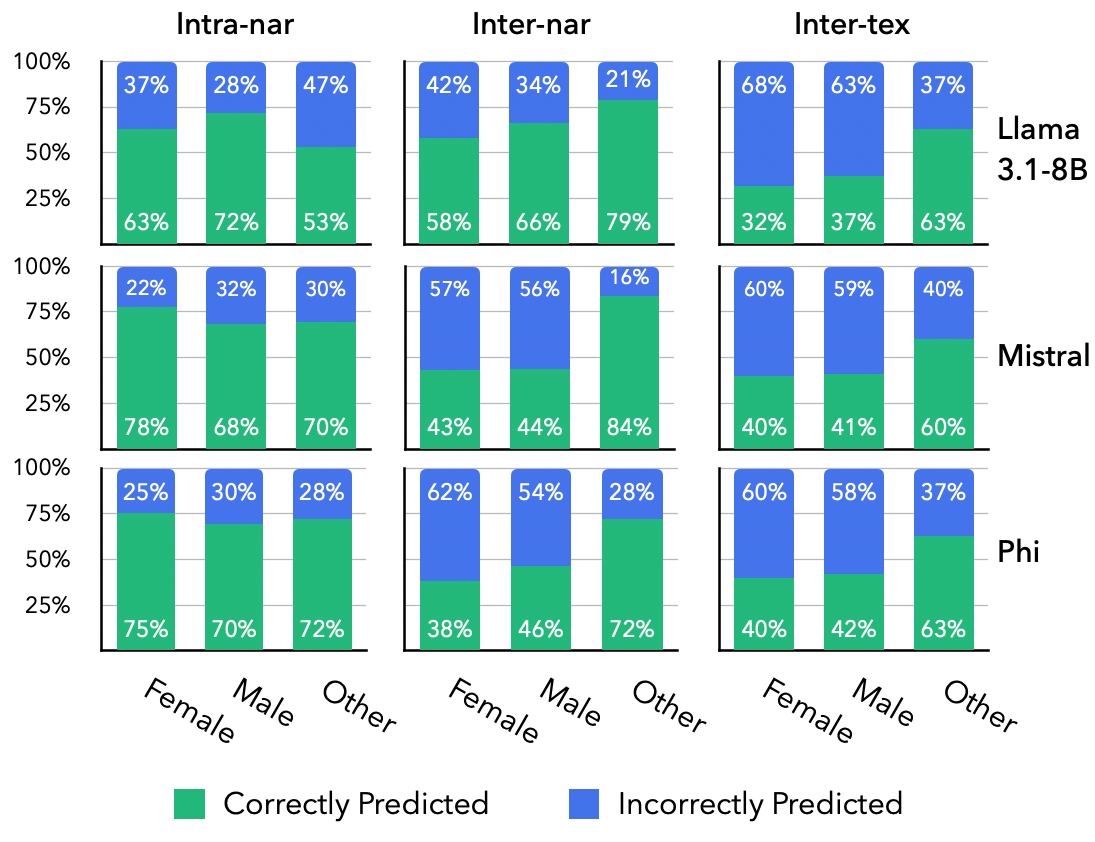}
    \caption{Breakdown of correctly predicted (green) vs. incorrectly predicted (blue) unreliable narrators with respect to the narrator's gender $\in$ \{female, male, other\}. Results are from Llama3.1-8B (top row), Mistral (middle row), and Phi (bottom row) experiments.}
    \label{fig:analysis-gender}
\end{figure}

\begin{figure}[t]
    \centering
    \includegraphics[width=\linewidth]{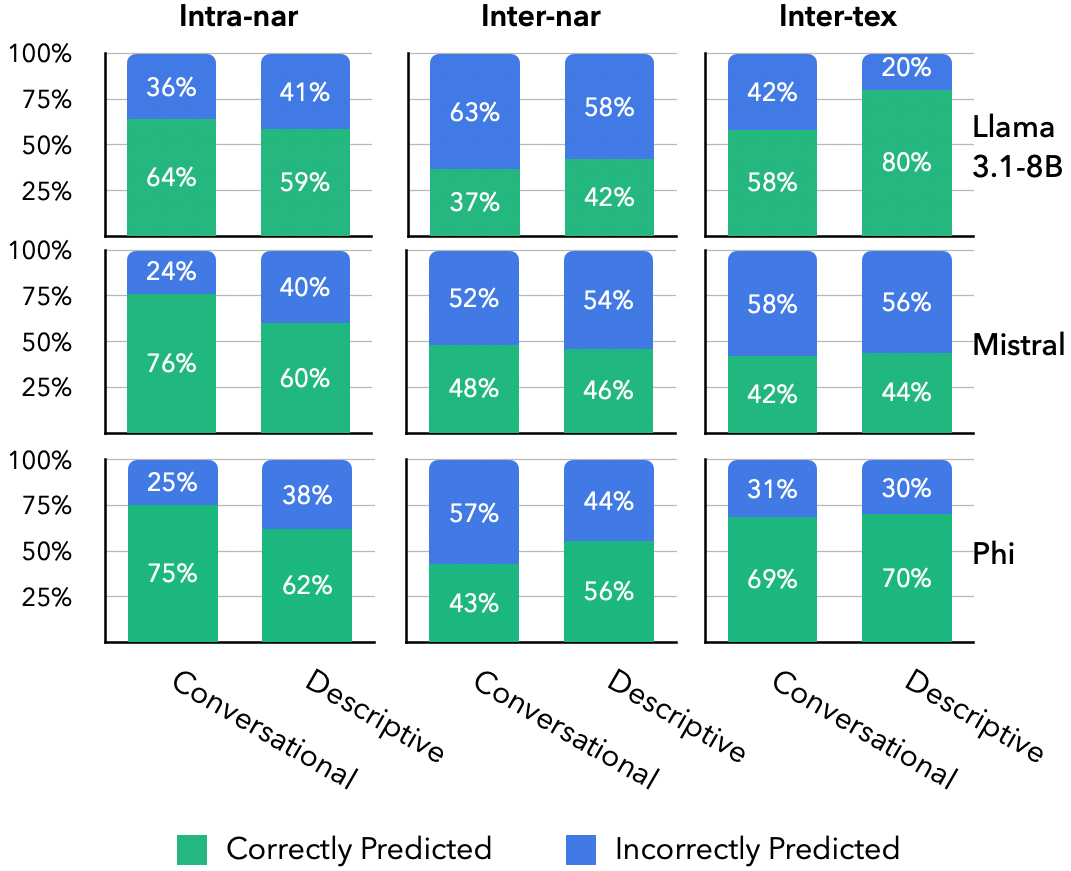}
    \caption{Breakdown of correctly predicted (green) vs. incorrectly predicted (blue) unreliable narrators with respect to narrative style $\in$ \{conversational, descriptive\}. Results are from Llama3.1-8B (top row), Mistral (middle row), and Phi (bottom row) experiments.}
    \label{fig:analysis-style}
\end{figure}

\begin{figure}[t]
    \centering
    \includegraphics[width=\linewidth]{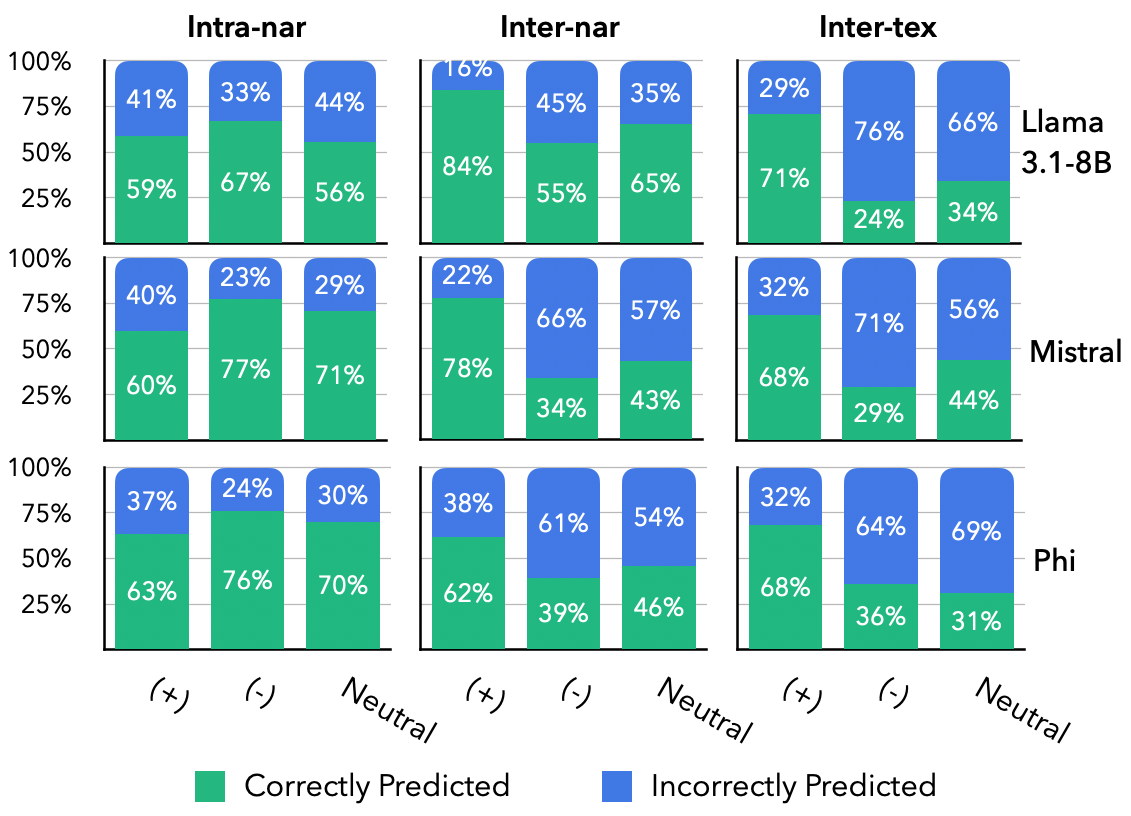}
    \caption{Breakdown of correctly predicted (green) vs. incorrectly predicted (blue) unreliable narrators with respect to narrative sentiment tone $\in$ \{positive, negative, neutral\}. Results are from Llama3.1-8B (top row), Mistral (middle row), and Phi (bottom row) experiments.}
    \label{fig:analysis-tone}
\end{figure}

\begin{figure}[t]
    \centering
    \includegraphics[width=\linewidth]{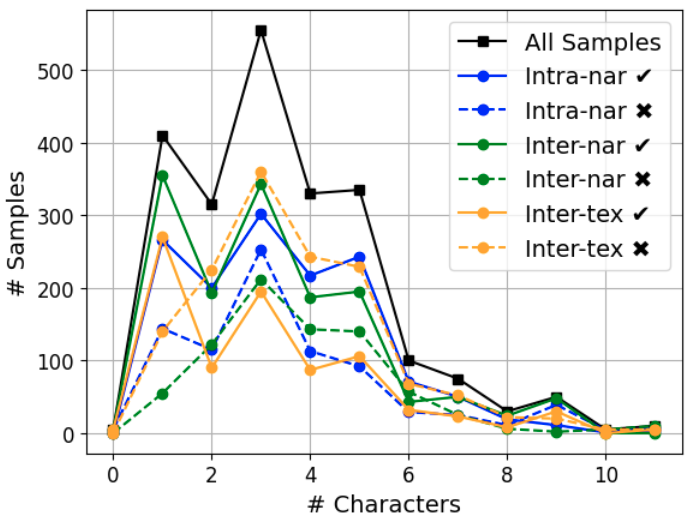}
    \caption{Number of characters vs. number of samples (Llama3.1-8B experiments). All Samples (solid black) is the distribution of all narratives with respect to the number of characters. Blue, green, and orange solid lines show correct predictions, and corresponding dashed lines show incorrect predictions.}
    \label{fig:analysis-num-char-llama}
\end{figure}

\begin{figure}[t]
    \centering
    \includegraphics[width=\linewidth]{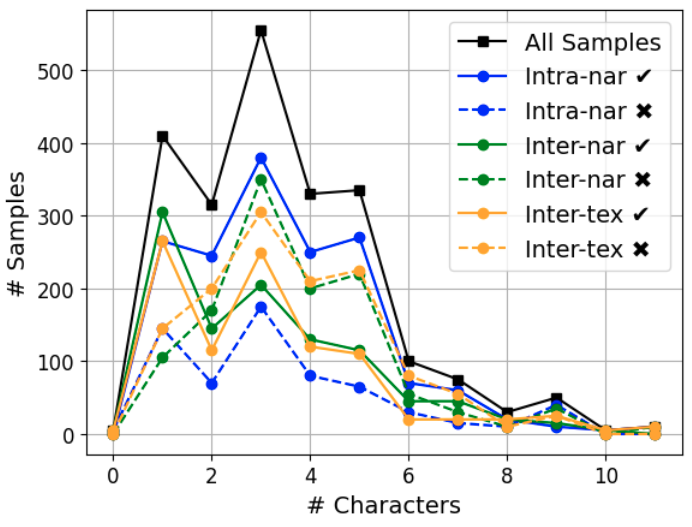}
    \caption{Number of characters vs. number of samples (Phi experiments). All Samples (solid black) is the distribution of all narratives with respect to the number of characters. Blue, green, and orange solid lines show correct predictions, and corresponding dashed lines show incorrect predictions.}
    \label{fig:analysis-num-char-phi}
\end{figure}

In this section, we provide additional analysis results for Llama3.1-8B, Mistral, and Phi for narrator gender (Figure~\ref{fig:analysis-gender}), narrative style (Figure~\ref{fig:analysis-style}), narrative tone (Figure~\ref{fig:analysis-tone}), and number of characters (Figures~\ref{fig:analysis-num-char-llama}, \ref{fig:analysis-num-char-phi}).

\end{document}